\documentclass[nonatbib]{article}
\usepackage{enumitem,kantlipsum}
\usepackage[preprint]{neurips_2024}

\usepackage[round]{natbib}
\renewcommand{\bibname}{References}
\renewcommand{\bibsection}{\subsubsection*{\bibname}}

\newtheorem{theorem}{Theorem}
\usepackage{graphicx} 
\usepackage{subcaption}
\usepackage{amsmath}
\usepackage{amssymb} 
\usepackage[dvipsnames]{xcolor}
\usepackage{algorithm}
\usepackage{algpseudocode}
\usepackage{booktabs} 
\usepackage{hyperref}
\usepackage{multirow}
\definecolor{bestblue}{RGB}{0,114,178}
\definecolor{secondorange}{RGB}{213,94,0}

\makeatletter
\newenvironment{breakablealgorithm}
  {
   \begin{center}
   \refstepcounter{algorithm}
   \hrule height .8pt depth 0pt \kern 2pt
   \renewcommand{\caption}[2][\relax]{%
     {\raggedright
      \textbf{\ALG@name~\thealgorithm} ##2\par}%
     \ifx\relax##1\relax
       \addcontentsline{loa}{algorithm}
       {\protect\numberline{\thealgorithm}##2}%
     \else
       \addcontentsline{loa}{algorithm}
       {\protect\numberline{\thealgorithm}##1}%
     \fi
     \kern 2pt
     \hrule
     \kern 2pt
   }%
  }
  {
   \kern 2pt
   \hrule
   \end{center}
  }
\makeatother

\usepackage[textsize=tiny]{todonotes}

\title{Beyond Distributional Fidelity: Causal-Penalized Diffusion for Synthetic Tabular Data}

\author{
  \textbf{Lan Tao}$^{1}$ \\
  \texttt{lantao@ucla.edu}
  \And
  \textbf{Yongxian He }$^{2}$ \\
  \texttt{yoh017@ucsd.edu}
  \And
  \textbf{Shirong Xu}$^{3}$ \\
  \texttt{shirongxu5566@xmu.edu.cn}
  \And
  \textbf{Yidong Ouyang}$^{1}$ \\
  \texttt{yidongouyang@g.ucla.edu}
  \And
  \textbf{Guang Cheng}$^{1}$ \\
  \texttt{guangcheng@ucla.edu}\\
  \and
  $^{1}$ Department of Statistics \& Data Science, University of California, Los Angeles \\
  $^{2}$ Herbert Wertheim School of Public Health and Human Longevity Science,\\ University of California, San Diego \\
  $^{3}$ Wang Yanan Institute for Studies in Economics, Xiamen University \\
}

\begin{document}

\maketitle

\begin{abstract}
Synthetic tabular generators are commonly optimized for distributional fidelity, but statistical similarity alone does not guarantee preservation of causal effects. In this paper, we study whether causal fidelity can be improved directly within a fully generative tabular model. \textit{Causal Fidelity} is defined with respect to a target estimand as the discrepancy between inferential distributions obtained from real and synthetic data, and theoretical results show that high statistical fidelity does not generally imply high causal fidelity. We then propose a causal-fidelity-aware training framework which adds a causal discrepancy penalty to the generative objective. The framework is instantiated with a causal-penalized TabDDPM and optimized using an on-policy score-function estimator. We further establish conditions under which causal regularization improves expected causal fidelity. Experiments across diverse treatment-effect simulations and two benchmark datasets evaluate the ability of our method to improve causal fidelity while preserving competitive statistical fidelity.
\end{abstract}


\section{Introduction}
Synthetic tabular data have become increasingly important for data sharing, privacy protection, data augmentation, and downstream machine learning. Recent generative models, including generative adversarial networks, variational autoencoders, and diffusion-based models, have substantially improved the realism of synthetic tabular data. Accordingly, the quality of synthetic data is commonly assessed through distributional fidelity, measuring how closely the generated data reproduce the statistical properties of the original population \citep{jordon2022synthetic}. Existing evaluation criteria range from feature-wise and pairwise discrepancies, to multivariate distributional distances \citep{herurkar2025evaluating, stoian2025survey,hernandez2025comprehensive}. While these criteria provide useful information about statistical similarity, they do not necessarily determine whether synthetic data preserve the information required for downstream causal inference. 

\begin{figure*}[t]
    \centering
    \includegraphics[width=\textwidth]
    {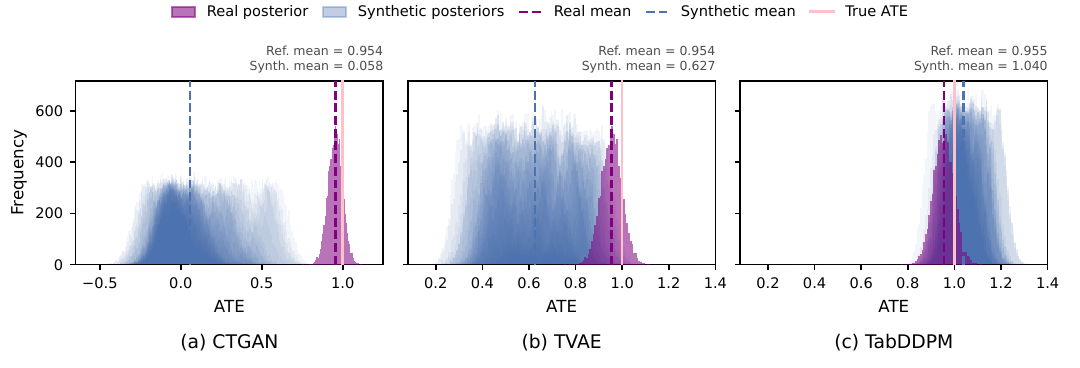}
    \caption{ATE posterior preservation for CTGAN, TVAE, and TabDDPM. For each generator, 10 independent models are trained on the same reference dataset and 10 synthetic datasets are sampled from each model, yielding 100 synthetic datasets per generator. CausalPFN is used to infer the ATE posterior from both the reference and synthetic datasets. Red denotes the reference posterior and blue denotes the synthetic-data posteriors. 
    }
    \label{fig:baseline-comparison-1}
\end{figure*}

Recent work has begun to examine whether synthetic tabular data preserve treatment-effect information beyond conventional statistical fidelity \citep{amad2026improving,xu2026generative}. To motivate this question empirically, we conduct a preliminary comparison of representative tabular generators. As shown in Figure~\ref{fig:baseline-comparison-1}, their ability to preserve the average treatment effect (ATE) posterior varies substantially across model families. CTGAN exhibits severe distortion of the ATE posterior, TVAE reduces but does not eliminate the discrepancy, and TabDDPM provides the strongest baseline while still showing a systematic shift from the reference posterior. These results suggest that causal information is not reliably preserved by standard tabular generators and motivate incorporating causal fidelity directly into the generative objective.

Existing approaches mainly address this issue through causal-aware evaluation and structured generation \citep{amad2026improving}, or hybrid data-generation strategies \citep{xu2026generative}. In contrast, we ask whether causal fidelity can be improved directly within the training objective of a fully generative tabular model. 

To this end, we first formalize \emph{causal fidelity} with respect to a target causal estimand as the discrepancy between inferential distributions obtained from real and synthetic data. We then propose a causal-fidelity-aware training framework that augments the standard generative objective with a causal discrepancy penalty. Because this discrepancy need not be differentiable with respect to the generator parameters, we optimize it using a score-function estimator, enabling a broad class of causal inference operators to be incorporated without requiring end-to-end differentiability.

\textbf{Contributions.} Our main contributions are summarized as follows:
\begin{itemize}
    \item We formalize \emph{causal fidelity} as the discrepancy between inferential distributions of a target causal estimand from real and synthetic data, and theoretically show that high distributional fidelity does not generally guarantee ATE preservation.
    \item We propose a causal-fidelity-aware training framework, which adds a causal discrepancy penalty to the generative objective, and instantiate it as an on-policy causal-penalized TabDDPM using a score-function estimator.
    \item We evaluate the proposed method across diverse treatment-effect simulations and two real-world benchmarks, showing that causal regularization improves causal fidelity while maintaining competitive distributional fidelity.
\end{itemize}

\section{Related Works}
In the following, we review three strands of literature most relevant to our study: (1) generative models for synthetic tabular data, (2) synthetic data for causal inference, and (3) causal effect estimation.

\subsection{Generative Models for Synthetic Tabular Data}
A wide range of deep generative models has been developed for synthetic tabular data. Earlier work includes GAN-based models such as CTGAN \citep{xu2019modeling} and TableGAN \citep{park2018data}, together with VAE-based approaches such as TVAE \citep{xu2019modeling} and RTVAE \citep{akrami2020robust}. Autoregressive language-model-based generators have also been explored for tabular synthesis, such as GReaT \citep{borisov2022language}. More recently, diffusion-based models have emerged as a competitive paradigm for tabular synthesis, including TabDDPM \citep{kotelnikov2023tabddpm} and TabSyn \citep{zhang2024mixed}. 

Another line of work incorporates task-specific feedback into tabular generation. ReTabSyn \citep{lin2026retabsyn} introduces a reinforcement-learning-based tabular synthesis framework to preserve feature correlations and predictive signals, whereas our method directly adds a causal discrepancy penalty to the generative objective to improve causal fidelity.

\subsection{Synthetic Data for Causal Inference}
Recent work has begun to examine whether synthetic tabular data preserve treatment-effect information beyond conventional statistical fidelity. \cite{amad2026improving} are among the first to systematically study synthetic tabular data for downstream treatment-effect analysis, and propose STEAM, a causal-aware framework for generating synthetic data in medical treatment-effect settings. \citet{xu2026generative} demonstrates that high distributional fidelity can coexist with distorted ATE estimates, and propose a hybrid synthesis framework which separates covariate generation from treatment and outcome modeling. CausalWrap \citep{asiaee2026causalwrap} applies a model-agnostic post-hoc correction using differentiable causal constraint functionals to improve structural and treatment-effect fidelity. In contrast, our work uses a score-function estimator, allowing non-differentiable causal discrepancy measures to be incorporated directly into generator training.

\subsection{Causal Effect Estimation}

Classical methods for causal effect estimation include outcome regression or g-computation \citep{robins1986new}, inverse probability weighting (IPW) \citep{horvitz1952generalization}, augmented IPW (AIPW) \citep{robins1994estimation}, and targeted maximum likelihood estimation (TMLE) \citep{van2006targeted}. Bayesian tree-based approaches further provide posterior uncertainty over treatment effects, including BART for causal inference \citep{hill2011bayesian} and Bayesian Causal Forests \citep{hahn2020bayesian}. More recently, amortized causal inference and causal foundation models have enabled fast, uncertainty-aware inference over causal quantities, including CausalPFN \citep{balazadeh2025causalpfn}, Do-PFN \citep{robertson2026pfn}, and CausalFM \citep{ma2026foundation}. Such methods are well aligned with our causal-fidelity-aware training framework, which measures fidelity through discrepancies between inferential distributions from real and synthetic data. In our experiments, we instantiate the causal inference operator using CausalPFN.

\section{Problem Setup and Modeling}
\label{sec: modeling}

\subsection{Preliminaries}
We adopt the potential outcomes framework for causal inference \citep{imbens2015causal}. Let $X \in \mathcal{X}$ denote covariates, $W \in \{0,1\}$ a binary treatment, and $Y \in \mathcal{Y}$ the observed outcome. For each unit, define potential outcomes $Y(0)$ and $Y(1)$ corresponding to the outcomes under control and treatment, respectively. However, only one of these is observed for each unit, yielding the fundamental problem of causal inference. The individual treatment effect is defined as $\tau_i = Y_i(1) - Y_i(0)$, which is generally unidentifiable. Consequently, inference focuses on population-level quantities. For example, the \textit{Average Treatment Effect (ATE)},
$$
\tau := \mathbb{E}[Y(1) - Y(0)],
$$
and the \textit{Conditional Average Treatment Effect (CATE)}, 
$$
\tau(x) := \mathbb{E}[Y(1) - Y(0) \mid X=x].
$$

Identification from observational data relies on standard assumptions: (i) consistency $Y = Y(W)$, (ii) unconfoundedness $(Y(0), Y(1)) \perp W \mid X$, and (iii) overlap $0 < \mathbb{P}(W=1 \mid X=x) < 1$. Under these assumptions, the ATE is identified from the observed-data
distribution through
\[
\tau
=
\mathbb{E}_{X}
\left[
\mathbb{E}(Y\mid W=1,X)
-
\mathbb{E}(Y\mid W=0,X)
\right].
\]

\subsection{Problem Statement} Let $D_{\text{real}} = \{(X_i, W_i, Y_i)\}_{i=1}^n$ denote an i.i.d.\ dataset drawn from an unknown distribution $\mathcal{P}$ over covariates $X \in \mathcal{X}$, treatment $W \in \{0,1\}$, and outcome $Y \in \mathcal{Y}$. We consider a generative model $G_\theta$ parameterized by $\theta$, which is trained on $D_{\text{real}}$ to learn an approximation $\hat{\mathcal{P}}_\theta$ of the data-generating distribution. A synthetic dataset of size $n_{\mathrm{syn}}$ is generated as follows
\[
D_{\mathrm{syn}}
=
\{(\widetilde X_i,\widetilde W_i,\widetilde Y_i)\}_{i=1}^{n_{\mathrm{syn}}},
\qquad
(\widetilde X_i,\widetilde W_i,\widetilde Y_i)
\overset{\mathrm{i.i.d.}}{\sim}
\widehat{\mathcal P}_\theta.
\]

To characterize the causal information contained in a dataset, let
$\mathcal A$ denote a fixed causal inference operator that maps a dataset
$D$ to an inferential distribution over the target causal estimand:
\[
\mathcal A(D)
=
\widehat p_D^{\mathcal A}(\tau).
\]

We define \emph{Causal Fidelity} with respect to the target estimand
$\tau$ as the extent to which synthetic data preserve the inferential
information about $\tau$ contained in the real data. Specifically, let
\[
\widehat p_{\mathrm{real}}(\tau)
=
\mathcal A(D_{\mathrm{real}})
\]
and
\[
\widehat p_{\mathrm{syn}}(\tau)
=
\mathcal A(D_{\mathrm{syn}}).
\]
For a distributional discrepancy $d(\cdot,\cdot)$, such as the
Wasserstein distance, we define \textit{causal discrepancy} of a synthetic
dataset as
\[
\mathcal D_{\tau}
(D_{\mathrm{real}},D_{\mathrm{syn}})
=
d\!\left(
\widehat p_{\mathrm{real}}(\tau),
\widehat p_{\mathrm{syn}}(\tau)
\right).
\]
A smaller discrepancy corresponds to higher causal fidelity.

We quantify the overall causal fidelity loss of $G_\theta$ by the expected discrepancy across synthetic datasets:
\begin{align}
\mathcal L_{\mathrm{causal}}(\theta)
& =
\mathbb E_{D_{\mathrm{syn}}\sim
\widehat{\mathcal P}_\theta^{\,n_{\mathrm{syn}}}}
\left[
\mathcal D_{\tau}
(D_{\mathrm{real}},D_{\mathrm{syn}})
\right]
\notag\\
& = 
\mathbb E_{D_{\mathrm{syn}}\sim
\widehat{\mathcal P}_\theta^{\,n_{\mathrm{syn}}}}
\left[
d\!\left(
\widehat p_{\mathrm{real}}(\tau),
\widehat p_{\mathrm{syn}}(\tau)
\right)
\right]
\end{align}



\subsection{Causal-Fidelity-Aware Training}

Standard generative training seeks to minimize a generative loss
$\mathcal L_{\mathrm{gen}}(\theta)$ that encourages
$\widehat{\mathcal P}_\theta$ to approximate the observational
distribution. However, this objective does not explicitly require the
generated data to preserve the target causal estimand. We therefore
consider the causal-regularized objective
\[
\min_\theta
\quad
\mathcal L_{\mathrm{gen}}(\theta)
+
\lambda \mathcal L_{\mathrm{causal}}(\theta),
\]
where $\lambda \geq 0$ controls the tradeoff between conventional
generative fidelity and causal fidelity.

A direct optimization of $\mathcal L_{\mathrm{causal}}(\theta)$ is
challenging, because the causal discrepancy is a dataset-level quantity evaluated only after sampling a synthetic dataset from $G_\theta$ and applying a causal inference operator $\mathcal A$. In general, $\mathcal A$ may involve estimation procedures that are not differentiable with respect to the generator parameters. Moreover, the discrete sampling process prevents gradients from being propagated directly through the generated dataset. Thus, $\nabla_\theta
\mathcal L_{\mathrm{causal}}(\theta)$ cannot generally be obtained through standard backpropagation.

For notational convenience, let $p_\theta(D)$ denote the probability
density or mass assigned by the generator to a synthetic dataset $D$. Then
the causal fidelity loss can be written as
\begin{align}
\mathcal L_{\mathrm{causal}}(\theta)
=
\mathbb E_{D_{\mathrm{syn}}\sim p_\theta}
\left[
\mathcal D_{\tau}
(D_{\mathrm{real}},D_{\mathrm{syn}})
\right].
\end{align}

Although $\mathcal D_{\tau}$ need not be differentiable with respect to
$\theta$, the score-function identity yields
\begin{equation}
\nabla_\theta \mathcal L_{\mathrm{causal}}(\theta)
=
\mathbb E_{D_{\mathrm{syn}}\sim p_\theta}
\Big[
\mathcal D_\tau(D_{\mathrm{real}},D_{\mathrm{syn}})
\cdot
\nabla_\theta \log p_\theta(D_{\mathrm{syn}})
\Big].
\label{eq:score_function}
\end{equation}

Thus, the causal discrepancy acts as a dataset-level weight on the
generator score. Synthetic datasets with larger causal discrepancy
contribute more strongly to the causal gradient, encouraging the
generator, in aggregate, to shift probability mass away from regions
associated with poor causal fidelity.

In practice, at training iteration $k$, let $\theta_k$ denote the current generator
parameters. We draw $m$ synthetic datasets
\[
D_{\mathrm{syn}}^{(j)}
\sim
p_{\theta_k},
\qquad j=1,\ldots,m,
\]
and compute their causal discrepancies
\[
\mathcal D_j
:=
\mathcal D_{\tau}
\left(
D_{\mathrm{real}},
D_{\mathrm{syn}}^{(j)}
\right)
=
d\!\left(
\widehat p_{\mathrm{real}}(\tau),
\widehat p_{D_{\mathrm{syn}}^{(j)}}(\tau)
\right).
\]
The score-function gradient can then be approximated by
\begin{align}
\nabla_\theta
\mathcal L_{\mathrm{causal}}(\theta)
\big|_{\theta=\theta_k}
\approx
\frac{1}{m}
\sum_{j=1}^m
\mathcal D_j
\nabla_\theta
\log p_\theta
\left(D_{\mathrm{syn}}^{(j)}\right)
\bigg|_{\theta=\theta_k}.
\end{align}

\subsection{Algorithm}
\label{sec: alg}

We instantiate our framework with TabDDPM, which shows stronger treatment-effect preservation than GAN- and VAE-based generators in our motivating experiments (Figure ~\ref{fig:baseline-comparison-1}). However, its causal fidelity becomes less reliable in higher-dimensional and more complex causal settings (Appendix ~\ref{append: vanilla_tabddpm}). This motivates adding explicit causal regularization to the TabDDPM training objective.

For diffusion-based generators, directly working with the marginal likelihood of the terminal synthetic dataset can be inconvenient because it requires integrating over the intermediate reverse-diffusion states. Rather than approximating this marginal likelihood, we augment the sampling space to include the full reverse-diffusion trajectory. Let $Z \sim P_\theta$ denote the collection of stochastic reverse trajectories generated by the model, and let $D_{\mathrm{syn}} = h(Z)$ denote the corresponding terminal synthetic dataset. Since $D_{\mathrm{syn}}$ is a deterministic function of the sampled trajectory, the expected causal loss can be equivalently written as
\begin{align}
\mathcal L_{\mathrm{causal}}(\theta)
=
\mathbb E_{Z\sim P_\theta}
\left[
\mathcal D_\tau
\bigl(
D_{\mathrm{real}},
h(Z)
\bigr)
\right].
\end{align}

This reformulation does not change the objective; it simply expresses the same expectation over the augmented trajectory space rather than over the terminal-data marginal. Applying the score-function identity to $P_\theta(Z)$ then yields

\begin{equation}
\begin{aligned}
\nabla_\theta
\mathcal L_{\mathrm{causal}}(\theta)
&=
\mathbb E_{Z\sim P_\theta}
\Big[
\mathcal D_\tau
\bigl(
D_{\mathrm{real}},h(Z)
\bigr)
\cdot
\nabla_\theta
\log P_\theta(Z)
\Big].
\end{aligned}
\label{eq:trajectory_score}
\end{equation}

For a reverse diffusion process, $P_\theta(Z)$ factorizes over the reverse transitions, so its score can be computed from the transition probabilities along the sampled trajectory. 

At each causal-fidelity update, the current generator samples $m$ reverse-diffusion trajectories $Z_j \sim P_\theta$, producing synthetic datasets $D_j=h(Z_j)$. Each dataset is assigned a causal discrepancy $\mathcal D_j$ by comparing its inferred ATE distribution with that of the reference data. The causal gradient is then estimated as a discrepancy-weighted average of the trajectory scores,
\begin{align}
g_{\mathrm{causal}}
=
\frac{\lambda}{m}
\sum_{j=1}^{m}
\operatorname{StopGrad}(\mathcal D_j-b)
\nabla_\theta \log P_\theta(Z_j),
\end{align}
where $b$ is a moving-average baseline used to reduce variance. 

Algorithm~\ref{alg:on_policy} summarizes the proposed training procedure, which combines the standard TabDDPM diffusion gradient with periodic on-policy causal-fidelity updates.

\begin{breakablealgorithm}
\caption{On-Policy Causal-Penalized TabDDPM}
\label{alg:on_policy}
\begin{algorithmic}[1]

\Require $D_{\mathrm{real}}$, generator $G_\theta$, causal operator
$\mathcal A$, penalty weight $\lambda$, number of synthetic datasets $m$,
iterations $K$, baseline decay $\rho$, gradient cap $\kappa$,
learning rates $\{\eta_k\}_{k=1}^K$
\Ensure Trained $G_\theta$

\State $\widehat p_{\mathrm{real}}(\tau)
\gets \mathcal A(D_{\mathrm{real}})$
\State Initialize $\theta$ and $b\gets 0$

\For{$k=1,\ldots,K$}
    \State $g_{\mathrm{gen}}
    \gets \nabla_\theta \mathcal L_{\mathrm{gen}}(\theta)$

    \If{causal update}
        \For{$j=1,\ldots,m$}
            \State $Z_j\sim P_\theta,\quad D_j=h(Z_j)$
            \State $\widehat p_j(\tau)\gets\mathcal A(D_j)$
            \State $\mathcal D_j
            \gets W_1(\widehat p_{\mathrm{real}}(\tau),
                       \widehat p_j(\tau))$
            \State $s_j\gets\nabla_\theta\log P_\theta(Z_j)$
        \EndFor

        \State $g_{\mathrm{causal}}
        \gets
        \frac{\lambda}{m}
        \sum_{j=1}^{m}
        \operatorname{StopGrad}(\mathcal D_j-b)s_j$

        \State $b\gets
        (1-\rho)b+\frac{\rho}{m}\sum_{j=1}^{m}\mathcal D_j$

        \State $\alpha\gets
        \min\!\left(
        1,\,
        \kappa
        \frac{\|g_{\mathrm{gen}}\|_2}
             {\|g_{\mathrm{causal}}\|_2+\varepsilon}
        \right)$

        \State $g\gets
        g_{\mathrm{gen}}
        +\alpha g_{\mathrm{causal}}$
    \Else
        \State $g\gets g_{\mathrm{gen}}$
    \EndIf

    \State $\theta\gets\theta-\eta_k g$
\EndFor

\State \Return $G_\theta$

\end{algorithmic}
\end{breakablealgorithm}

\section{Theory}
\label{sec:theory}

In this section, we examine the relationship between joint distributional fidelity and ATE preservation. Under the problem setup, we show that joint distributional proximity alone does not uniformly control ATE error. We then establish a bound on the expected causal discrepancy of regularized solutions and a sufficient condition for strict improvement. 

\subsection{Joint Distributional Fidelity and the ATE}

For a synthetic distribution $\widehat{\mathcal P}_\theta$, we define its population ATE as
\[
\begin{aligned}
\widetilde\tau_\theta
=\mathbb E_{\widetilde X}\!\bigl[
\mathbb E(\widetilde Y\mid\widetilde W=1,\widetilde X)-\mathbb E(\widetilde Y\mid\widetilde W=0,\widetilde X)\bigr],
\end{aligned}
\]
where all expectations are taken under $\widehat{\mathcal P}_\theta$. The total variation distance is denoted by
$d_{\mathrm{TV}}(\mathcal P,\widehat{\mathcal P}_\theta)
=\sup_B|\mathcal P(B)-\widehat{\mathcal P}_\theta(B)|$,
where the supremum is over measurable sets $B$.

\begin{theorem}
\label{thm:joint-ate-gap}
Fix $\Delta>0$ and $\sigma>0$. For each $\epsilon\in(0,1)$,
consider the observational and generated distributions given by
\[
\begin{aligned}
\mathcal P:\quad
&X\sim\mathcal N(0,1),\quad
W\sim\operatorname{Bernoulli}(\epsilon),\\
&Y=U,\quad U\sim\mathcal N(0,\sigma^2),\\
\widehat{\mathcal P}_\theta:\quad
&\widetilde X\sim\mathcal N(0,1),\quad
\widetilde W\sim\operatorname{Bernoulli}(\epsilon),\\
&\widetilde Y=\Delta\widetilde W+\widetilde U,\quad
\widetilde U\sim\mathcal N(0,\sigma^2).
\end{aligned}
\]
Within each distribution, the covariate, treatment, and noise are mutually independent. The dependence of $\mathcal P$ and $\theta$ on $\epsilon$ is suppressed in the notation. Both distributions admit potential outcomes satisfying consistency, unconfoundedness, and the pointwise overlap condition in the problem setup. However,
\begin{align}
|\widetilde\tau_\theta-\tau|&=\Delta
\mbox{ and }
d_{\mathrm{TV}}(\mathcal P,\widehat{\mathcal P}_\theta)\leq\epsilon,
\label{eq:theory-tv}
\end{align}
As $\epsilon\rightarrow 0$, the total variation distance converges to zero, whereas the ATE discrepancy remains $\Delta$.
\end{theorem}

The construction isolates the impact of rare treatment assignments on joint distributional discrepancies. Since the real and generated distributions differ only in the treated group, their contribution to the joint TV distance is weighted by the treatment probability $\epsilon$. In contrast, the ATE depends directly on the treated conditional mean, so the ATE discrepancy can remain fixed even as the joint TV distance approaches zero. Thus, Theorem~\ref{thm:joint-ate-gap} rules out any uniform ATE error bound that vanishes with this joint distributional discrepancy under pointwise overlap alone. At the same time, exact equality of the joint distributions still implies equality of the corresponding identified ATEs.

\subsection{Effect of Causal Regularization}

Fix the reference dataset $D_{\mathrm{real}}$, the inference operator
$\mathcal A$, the synthetic sample size $n_{\mathrm{syn}}$, and a
nonempty generator class $\{G_\theta\mid\theta\in\Theta\}$.
We set $d=W_1$, as in Algorithm~\ref{alg:on_policy}, and retain the
losses $\mathcal L_{\mathrm{gen}}$ and $\mathcal L_{\mathrm{causal}}$
defined in the problem setup. Assume that the inferred ATE
distributions have finite first moments and that both losses are
finite on $\Theta$, with
\[
\mathcal L_{\mathrm{gen}}^\star
=\inf_{\theta\in\Theta}\mathcal L_{\mathrm{gen}}(\theta)>-\infty.
\]
Unless specified otherwise, expectations are taken over
$D_{\mathrm{syn}}\sim\widehat{\mathcal P}_\theta^{\,n_{\mathrm{syn}}}$
conditional on $D_{\mathrm{real}}$.

\begin{theorem}[Causal regularization]
\label{thm:causal-penalty}
Let $\lambda>0$ and $\delta_{\mathrm{opt}}\geq0$. Suppose
$\theta_\lambda\in\Theta$ satisfies the approximate optimality condition
\begin{equation}
\begin{aligned}
&\mathcal L_{\mathrm{gen}}(\theta_\lambda)
+\lambda\mathcal L_{\mathrm{causal}}(\theta_\lambda)\leq\inf_{\theta\in\Theta}\bigl\{
\mathcal L_{\mathrm{gen}}(\theta)
+\lambda\mathcal L_{\mathrm{causal}}(\theta)\bigr\}
+\delta_{\mathrm{opt}}.
\end{aligned}
\label{eq:theory-approx-opt}
\end{equation}
Then, for every $\bar\theta\in\Theta$,
\begin{equation}
\begin{aligned}
\mathcal L_{\mathrm{causal}}(\theta_\lambda)
&\leq\mathcal L_{\mathrm{causal}}(\bar\theta)+\frac{\mathcal L_{\mathrm{gen}}(\bar\theta)
-\mathcal L_{\mathrm{gen}}^\star+\delta_{\mathrm{opt}}}{\lambda}.
\end{aligned}
\label{eq:theory-oracle}
\end{equation}
If an unpenalized minimizer
$\theta_{\mathrm{gen}}\in\arg\min_{\theta\in\Theta}
\mathcal L_{\mathrm{gen}}(\theta)$
exists, then
\[
\mathcal L_{\mathrm{causal}}(\theta_\lambda)
\leq\mathcal L_{\mathrm{causal}}(\theta_{\mathrm{gen}})
+\delta_{\mathrm{opt}}/\lambda.
\]
For any such $\theta_{\mathrm{gen}}$, if a comparator
$\bar\theta\in\Theta$ satisfies
$\mathcal L_{\mathrm{causal}}(\bar\theta)
<\mathcal L_{\mathrm{causal}}(\theta_{\mathrm{gen}})$,
then $\mathcal L_{\mathrm{causal}}(\theta_\lambda)
<\mathcal L_{\mathrm{causal}}(\theta_{\mathrm{gen}})$ whenever
\begin{equation}
\lambda>
\frac{\mathcal L_{\mathrm{gen}}(\bar\theta)
-\mathcal L_{\mathrm{gen}}^\star+\delta_{\mathrm{opt}}}
{\mathcal L_{\mathrm{causal}}(\theta_{\mathrm{gen}})
-\mathcal L_{\mathrm{causal}}(\bar\theta)}.
\label{eq:theory-strict-improvement}
\end{equation}
For exact global minimizers $\theta_{\lambda_i}\in\Theta$ of
$\mathcal L_{\mathrm{gen}}+\lambda_i\mathcal L_{\mathrm{causal}}$
at $0\leq\lambda_1<\lambda_2$, respectively,
\begin{align}
\mathcal L_{\mathrm{causal}}(\theta_{\lambda_2})
&\leq\mathcal L_{\mathrm{causal}}(\theta_{\lambda_1}),
\label{eq:theory-causal-monotone}\\
\mathcal L_{\mathrm{gen}}(\theta_{\lambda_2})
&\geq\mathcal L_{\mathrm{gen}}(\theta_{\lambda_1}).
\label{eq:theory-generative-monotone}
\end{align}
\end{theorem}

Exact minimization of the regularized objective yields an expected
causal discrepancy no greater than that of an unpenalized global
minimizer. Equation~\eqref{eq:theory-strict-improvement} provides
a sufficient condition for strict improvement.
Equation~\eqref{eq:theory-oracle} further relates the causal loss
to that of any comparator, with the comparator's excess generative
loss and the optimization error scaled by $1/\lambda$.
Along exact global optima, the causal loss is nonincreasing in
$\lambda$, while the generative loss is nondecreasing.

\textbf{Example (Gaussian model with an omitted intercept).}
Consider the Gaussian models
\[
\begin{aligned}
\mathcal P:&\quad Y=\beta_0+\tau W+U,\\
\widehat{\mathcal P}_\theta:&\quad
\widetilde Y=\theta\widetilde W+\widetilde U,
\end{aligned}
\]
where $\beta_0\ne0$, $\tau,\theta\in\mathbb R$, and
$W,\widetilde W\sim\operatorname{Bernoulli}(1/2)$.
The noise terms follow $\mathcal N(0,\sigma^2)$ for fixed
$\sigma>0$ and are independent of their respective treatment
indicators. Covariates have the same independent distribution under
both models. With $\mathcal L_{\mathrm{gen}}(\theta)
=D_{\mathrm{KL}}(\mathcal P\Vert\widehat{\mathcal P}_\theta)$,
the unpenalized minimizer is $\theta_{\mathrm{gen}}=\tau+\beta_0$,
while the generated ATE is $\widetilde\tau_\theta=\theta$.
Consider the population penalty
$W_1(\delta_\tau,\delta_{\widetilde\tau_\theta})=|\theta-\tau|$,
where $\delta_a$ denotes a point mass at $a$.
For $\lambda\geq0$, the unique minimizer of
$\mathcal L_{\mathrm{gen}}(\theta)+\lambda|\theta-\tau|$ is
\begin{equation}
\theta_\lambda=\tau+\operatorname{sgn}(\beta_0)
\bigl(|\beta_0|-2\lambda\sigma^2\bigr)_+,
\label{eq:theory-gaussian-solution}
\end{equation}
where $(a)_+=\max\{a,0\}$. Hence
\[
|\widetilde\tau_{\theta_\lambda}-\tau|
=\bigl(|\beta_0|-2\lambda\sigma^2\bigr)_+.
\]
Thus every $\lambda>0$ strictly reduces the ATE error relative to
the unpenalized solution. The generated ATE equals $\tau$ whenever
$\lambda\geq|\beta_0|/(2\sigma^2)$.
Appendix~\ref{app:theory-gaussian-example} provides the derivation
and relates this population penalty to its finite-sample counterpart.

\textbf{Implications for ATE estimation.}
Assume $\tau$ is finite. Define the means of the inferred ATE
distributions as
\[
\widehat\tau_{\mathrm{real}}
=\mathbb E_{t\sim\widehat p_{\mathrm{real}}}[t],
\qquad
\widehat\tau_{\mathrm{syn}}
=\mathbb E_{t\sim\widehat p_{\mathrm{syn}}}[t].
\]
The estimator $\widehat\tau_{\mathrm{syn}}$ depends on the sampled
dataset $D_{\mathrm{syn}}$, while $\widetilde\tau_\theta$ is the
population ATE under $\widehat{\mathcal P}_\theta$.
Define the absolute reference estimation error as
$e_{\mathrm{real}}=|\widehat\tau_{\mathrm{real}}-\tau|$.
Then the expected absolute estimation error satisfies
\begin{equation}
\mathbb E\bigl|\widehat\tau_{\mathrm{syn}}-\tau\bigr|
\leq e_{\mathrm{real}}+\mathcal L_{\mathrm{causal}}(\theta).
\label{eq:theory-mean-error}
\end{equation}
Combining Equations~\eqref{eq:theory-mean-error}
and~\eqref{eq:theory-oracle} gives an estimation error bound for
the regularized generator. The bound separates the reference
estimation error from the inferential discrepancy introduced by
synthesis. Reducing $\mathcal L_{\mathrm{causal}}$ tightens this
bound without implying a monotone reduction in the actual estimation
error. Appendix~\ref{app:theory-ate-transfer} gives a corresponding
bound for $|\widetilde\tau_\theta-\tau|$ that also accounts for
estimation error under the generated distribution.

\section{Experiments}
\label{sec:experiments}

In this section, we present the experimental results. First, we conduct simulation studies to evaluate whether incorporating a causal-fidelity penalty improves the preservation of causal effects in synthetic data. Second, we compare the proposed causal-penalized TabDDPM with several baseline tabular data generators on two real-world datasets, evaluating both statistical fidelity and causal fidelity.

\subsection{Simulation Study}

\textbf{Experimental Setup.} We compared baseline TabDDPM and causal-penalized TabDDPM in three selected scenarios: (i) linear outcome models with a constant treatment effect, (ii) nonlinear outcome models with a constant treatment effect, and (iii) heterogeneous treatment-effect models.

Each scenario included $25$ continuous and five categorical covariates, randomized binary treatment, and a continuous outcome with Gaussian noise. Both methods were trained on $2,000$ observations for $20,000$ iterations using the same architecture and optimization settings. The causal penalty matched synthetic-data and reference-data ATE posteriors inferred by CausalPFN. Each fitted model generated $20$ synthetic datasets of $3,000$ observations.

\textbf{Evaluation Metrics.} We evaluate both distributional and causal fidelity of each synthetic dataset. Distributional fidelity is measured by mean Wasserstein-1 distance for continuous covariates, mean Jensen–Shannon distance for categorical covariates, total variation distance for treatment, Wasserstein-1 distance for outcome, and correlation discrepancy. Causal fidelity is assessed by comparing the ATE posteriors inferred from the reference and synthetic datasets using Wasserstein-1 distance, absolute difference in posterior means, and RMSE relative to the ground-truth effect value.

\textbf{Results.} 
As shown in Table~\ref{tab:simulation_results}, causal-penalized TabDDPM consistently improves causal fidelity across all three simulation settings compared with vanilla TabDDPM. In the linear setting, it also improves all distributional-fidelity metrics. In the nonlinear and heterogeneous settings, the gains in causal fidelity come with modest degradation in some distributional metrics. These results demonstrate that the proposed causal penalty effectively improves preservation of treatment effects across increasingly complex causal mechanisms. 

\begin{figure*}[h]
    \centering
    \centering
    \includegraphics[width=\linewidth]{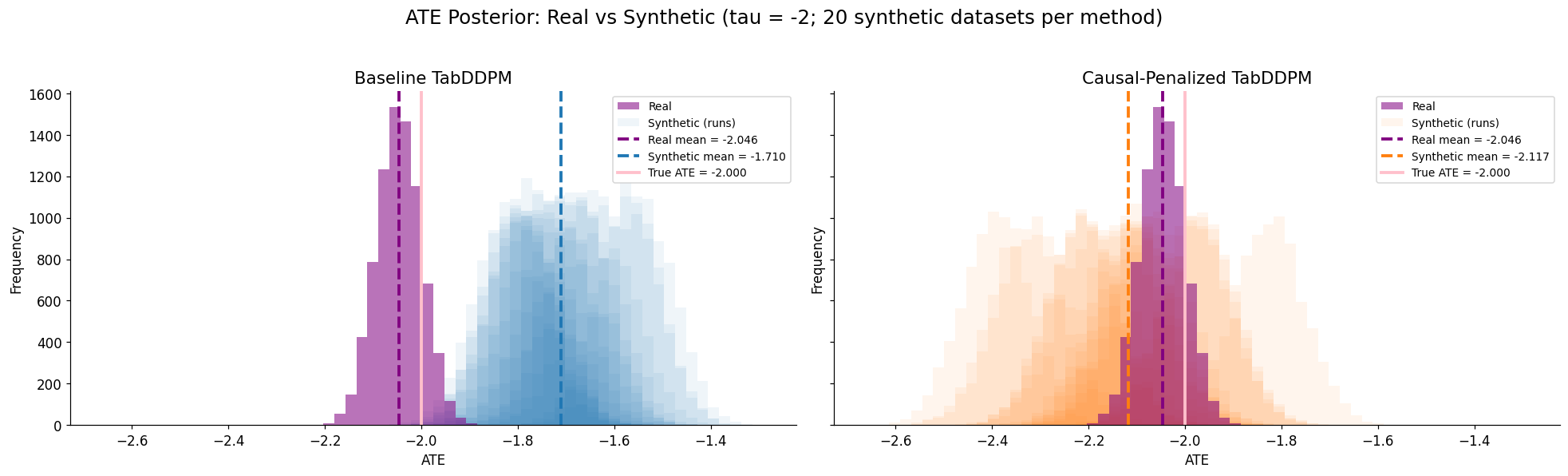}
    \caption{ATE posterior distributions estimated from real and synthetic data in the linear simulation setting with true ATE $\tau=-2$. Synthetic datasets are generated using vanilla TabDDPM (left) and the proposed causal-penalized TabDDPM (right). The causal-fidelity penalty shifts the synthetic-data posterior toward the posterior estimated from the real data and the true ATE.}
    \label{fig: simulation}
\end{figure*}

\begin{table*}[h]
\centering
\caption{
Comparison of vanilla TabDDPM and causal-penalized TabDDPM across three simulation settings: linear constant treatment effect ($\tau=-2$), nonlinear constant treatment effect ($\tau=1$), and heterogeneous treatment effects ($\tau=1.55$). 
}
\label{tab:simulation_results}

\resizebox{\textwidth}{!}{
\begin{tabular}{llcccccccc}
\toprule

&&
\multicolumn{5}{c}{\textit{Distributional fidelity}}
&
\multicolumn{3}{c}{\textit{Causal fidelity}}
\\

\cmidrule(lr){3-7}
\cmidrule(lr){8-10}

\textbf{Setting}
&
\textbf{Method}
&
\textbf{Num. W$_1$ $\downarrow$}
&
\textbf{Cat. JS $\downarrow$}
&
\textbf{T TVD $\downarrow$}
&
\textbf{Y W$_1$ $\downarrow$}
&
\textbf{Corr. L2 $\downarrow$}
&
\textbf{Post. W$_1$ $\downarrow$}
&
\textbf{Mean Gap $\downarrow$}
&
\textbf{ATE RMSE $\downarrow$}
\\

\midrule

\multirow{2}{*}{Linear}
&
TabDDPM
&
$0.877_{\pm 0.021}$
&
$0.065_{\pm 0.003}$
&
$0.016_{\pm 0.007}$
&
$1.899_{\pm 0.048}$
&
$5.115_{\pm 0.159}$
&
$0.336_{\pm 0.086}$
&
$0.336_{\pm 0.086}$
&
0.302
\\

&
Causal-TabDDPM
&
$\mathbf{0.737}_{\pm 0.017}$
&
$\mathbf{0.044}_{\pm 0.003}$
&
$\mathbf{0.006}_{\pm 0.005}$
&
$\mathbf{1.596}_{\pm 0.042}$
&
$\mathbf{4.967}_{\pm 0.140}$
&
$\mathbf{0.123}_{\pm 0.093}$
&
$\mathbf{0.122}_{\pm 0.094}$
&
\textbf{0.179}
\\

\midrule

\multirow{2}{*}{Nonlinear}
&
TabDDPM
&
$\mathbf{0.688}_{\pm 0.016}$
&
$0.043_{\pm 0.002}$
&
$\mathbf{0.068}_{\pm 0.010}$
&
$\mathbf{2.120}_{\pm 0.061}$
&
$\mathbf{3.604}_{\pm 0.095}$
&
$0.213_{\pm 0.107}$
&
$0.212_{\pm 0.108}$
&
0.233
\\

&
Causal-TabDDPM
&
$0.759_{\pm 0.020}$
&
$\mathbf{0.040}_{\pm 0.004}$
&
$0.086_{\pm 0.009}$
&
$2.254_{\pm 0.071}$
&
$6.404_{\pm 0.135}$
&
$\mathbf{0.066}_{\pm 0.060}$
&
$\mathbf{0.061}_{\pm 0.063}$
&
\textbf{0.086}
\\

\midrule

\multirow{2}{*}{Heterogeneous}
&
TabDDPM
&
$\mathbf{0.372}_{\pm 0.012}$
&
$\mathbf{0.024}_{\pm 0.002}$
&
$\mathbf{0.013}_{\pm 0.008}$
&
$\mathbf{0.780}_{\pm 0.035}$
&
$\mathbf{2.629}_{\pm 0.125}$
&
$0.125_{\pm 0.068}$
&
$0.124_{\pm 0.069}$
&
0.136
\\

&
Causal-TabDDPM
&
$0.392_{\pm 0.014}$
&
$0.033_{\pm 0.003}$
&
$0.018_{\pm 0.009}$
&
$0.830_{\pm 0.038}$
&
$3.070_{\pm 0.132}$
&
$\mathbf{0.061}_{\pm 0.049}$
&
$\mathbf{0.060}_{\pm 0.049}$
&
\textbf{0.081}
\\

\bottomrule
\end{tabular}
}
\end{table*}

\subsection{Real Application}

We further evaluate causal-penalized TabDDPM and other tabular data generators on two causal benchmark datasets, IHDP and IST, which preserve realistic covariate structures and treatment-assignment mechanisms.

\textbf{Datasets.}
We use two datasets that complement the controlled simulations from different perspectives. IHDP provides a standard semi-synthetic benchmark with known treatment effects, while IST allows us to evaluate the method using data from a large randomized clinical trial.

\begin{itemize}
    \item \textbf{IHDP dataset:}
    The Infant Health and Development Program (IHDP) is a widely used semi-synthetic benchmark for causal inference \citep{hill2011bayesian}. It contains 747 subjects with 25 baseline covariates describing characteristics of the children and their mothers. The treatment corresponds to an intensive early-childhood intervention that included specialist home visits and high-quality child care. The outcome is measured using later cognitive test scores. In the benchmark dataset, the covariates and treatment assignments come from the original study, while the potential outcomes are simulated.

    \item \textbf{IST dataset:}
    The International Stroke Trial (IST) is a large randomized controlled trial of antithrombotic treatment for acute ischaemic stroke \citep{ist1997}. We use randomized aspirin assignment (\texttt{RXASP}) as the binary treatment and recurrent ischaemic stroke within 14 days (\texttt{DRSISC}) as the outcome. Following the IST analysis in \citet{gruber2024targeted}, we retain the same set of 24 baseline analysis variables. These cover demographic characteristics, stroke presentation and severity, prior medication use, imaging findings, and other clinical information. After preprocessing, the final analysis dataset contains 19,408 patients.
\end{itemize}

\textbf{Baselines and Metrics.}
We compare our proposed method with four representative tabular data generators: (1) \textbf{CTGAN} \citep{xu2019modeling}, a conditional GAN designed for mixed-type tabular data; (2) \textbf{TVAE} \citep{xu2019modeling}, a variational autoencoder that models the joint distribution through a latent representation; (3) \textbf{TabSyn} \citep{zhang2024mixed}, a diffusion-based method that maps heterogeneous tabular features into a continuous latent space before performing diffusion; and (4) \textbf{TabDDPM} \citep{kotelnikov2023tabddpm}, a diffusion model designed directly for heterogeneous tabular data. We evaluate all methods in terms of both distributional and causal fidelity using the same metrics as in the simulation study.


\textbf{Results.}
The results on IHDP and IST are reported in Table~\ref{tab:real_application}. Relative to vanilla TabDDPM, causal regularization reduces the ATE posterior Wasserstein distance and posterior mean gap by $27.3\%$ and $30.4\%$ on IHDP, and by $22.7\%$ and $24.5\%$ on IST.

These improvements are accompanied by relatively small changes in conventional distributional fidelity. On IHDP, Causal-TabDDPM ranks second across all distributional metrics and substantially improves the causal metrics over vanilla TabDDPM, approaching the performance of TabSyn. On IST, it remains competitive on most distributional metrics and achieves the lowest ATE posterior Wasserstein distance and posterior mean gap. Overall, causal regularization consistently improves causal fidelity over vanilla TabDDPM while maintaining strong distributional fidelity across both datasets. Additional visualizations are provided in Appendix~\ref{append:application-results}.

\newcommand{\mstd}[2]{#1_{\scriptscriptstyle \pm #2}}

\begin{table*}[t]
\centering
\caption{
Comparison of synthetic tabular data generators on the IHDP and IST datasets with respect to distributional and causal fidelity. The best and second-best results for each metric are highlighted in \textcolor{bestblue}{\textbf{blue}} and \textcolor{secondorange}{\textbf{orange}}, respectively.
}
\label{tab:real_application}

\scriptsize
\setlength{\tabcolsep}{3.5pt}

\resizebox{\textwidth}{!}{%
\begin{tabular}{@{}lccccccc@{}}
\toprule
& \multicolumn{5}{c}{\textit{Distributional fidelity}}
& \multicolumn{2}{c}{\textit{Causal fidelity}} \\
\cmidrule(lr){2-6}
\cmidrule(lr){7-8}

Method
& Num. WD $\downarrow$
& Cat. JS $\downarrow$
& T TVD $\downarrow$
& Y $W_1$ $\downarrow$
& Corr. $L_2$ $\downarrow$
& Post. $W_1$ $\downarrow$
& Mean Gap $\downarrow$ \\
\midrule

\multicolumn{8}{@{}l}{\textbf{IHDP dataset}} \\[1pt]

CTGAN
& $\mstd{0.092}{0.001}$
& $\mstd{0.023}{0.001}$
& $\mstd{0.015}{0.006}$
& $\mstd{0.931}{0.031}$
& $\mstd{3.056}{0.014}$
& $\mstd{3.928}{0.131}$
& $\mstd{3.928}{0.131}$ \\

TVAE
& $\mstd{0.035}{0.001}$
& $\mstd{0.159}{0.001}$
& $\mstd{0.107}{0.005}$
& $\mstd{0.464}{0.029}$
& $\mstd{3.060}{0.101}$
& $\mstd{0.110}{0.065}$
& $\mstd{0.107}{0.069}$ \\

TabSyn
& $\mstd{0.027}{0.001}$
& $\mstd{0.028}{0.002}$
& $\mstd{0.016}{0.007}$
& $\mstd{0.290}{0.027}$
& $\mstd{0.486}{0.025}$
& $\textcolor{bestblue}{\mathbf{0.056}}_{\scriptscriptstyle \pm 0.022}$
& $\textcolor{bestblue}{\mathbf{0.043}}_{\scriptscriptstyle \pm 0.030}$ \\

TabDDPM
& $\textcolor{bestblue}{\mathbf{0.010}}_{\scriptscriptstyle \pm 0.001}$
& $\textcolor{bestblue}{\mathbf{0.011}}_{\scriptscriptstyle \pm 0.001}$
& $\textcolor{bestblue}{\mathbf{0.013}}_{\scriptscriptstyle \pm 0.007}$
& $\textcolor{bestblue}{\mathbf{0.118}}_{\scriptscriptstyle \pm 0.027}$
& $\textcolor{bestblue}{\mathbf{0.430}}_{\scriptscriptstyle \pm 0.051}$
& $\mstd{0.133}{0.059}$
& $\mstd{0.131}{0.060}$ \\

Causal-TabDDPM
& $\textcolor{secondorange}{\mathbf{0.011}}_{\scriptscriptstyle \pm 0.001}$
& $\textcolor{secondorange}{\mathbf{0.012}}_{\scriptscriptstyle \pm 0.002}$
& $\textcolor{secondorange}{\mathbf{0.014}}_{\scriptscriptstyle \pm 0.007}$
& $\textcolor{secondorange}{\mathbf{0.126}}_{\scriptscriptstyle \pm 0.029}$
& $\textcolor{secondorange}{\mathbf{0.466}}_{\scriptscriptstyle \pm 0.039}$
& $\textcolor{secondorange}{\mathbf{0.097}}_{\scriptscriptstyle \pm 0.048}$
& $\textcolor{secondorange}{\mathbf{0.092}}_{\scriptscriptstyle \pm 0.053}$ \\

\midrule

\multicolumn{8}{@{}l}{\textbf{IST dataset}} \\[1pt]

CTGAN
& $\mstd{0.027}{0.001}$
& $\mstd{0.062}{0.001}$
& $\mstd{0.039}{0.007}$
& $\mstd{0.013}{0.002}$
& $\mstd{0.970}{0.018}$
& $\mstd{0.012}{0.006}$
& $\mstd{0.012}{0.006}$ \\

TVAE
& $\mstd{0.021}{0.002}$
& $\mstd{0.041}{0.002}$
& $\mstd{0.071}{0.009}$
& $\mstd{0.022}{0.000}$
& $\mstd{1.392}{0.123}$
& $\mstd{0.011}{0.000}$
& $\mstd{0.011}{0.000}$ \\

TabSyn
& $\mstd{0.005}{0.001}$
& $\textcolor{bestblue}{\mathbf{0.011}}_{\scriptscriptstyle \pm 0.002}$
& $\textcolor{bestblue}{\mathbf{0.007}}_{\scriptscriptstyle \pm 0.005}$
& $\textcolor{bestblue}{\mathbf{0.002}}_{\scriptscriptstyle \pm 0.002}$
& $\textcolor{bestblue}{\mathbf{0.268}}_{\scriptscriptstyle \pm 0.020}$
& $\textcolor{secondorange}{\mathbf{0.005}}_{\scriptscriptstyle \pm 0.002}$
& $\textcolor{secondorange}{\mathbf{0.005}}_{\scriptscriptstyle \pm 0.002}$ \\

TabDDPM
& $\textcolor{bestblue}{\mathbf{0.004}}_{\scriptscriptstyle \pm 0.001}$
& $\mstd{0.016}{0.002}$
& $\mstd{0.010}{0.007}$
& $\mstd{0.004}{0.002}$
& $\textcolor{secondorange}{\mathbf{0.301}}_{\scriptscriptstyle \pm 0.024}$
& $\mstd{0.006}{0.001}$
& $\mstd{0.006}{0.001}$ \\

Causal-TabDDPM
& $\textcolor{bestblue}{\mathbf{0.004}}_{\scriptscriptstyle \pm 0.001}$
& $\textcolor{secondorange}{\mathbf{0.016}}_{\scriptscriptstyle \pm 0.002}$
& $\textcolor{secondorange}{\mathbf{0.009}}_{\scriptscriptstyle \pm 0.007}$
& $\textcolor{secondorange}{\mathbf{0.003}}_{\scriptscriptstyle \pm 0.002}$
& $\mstd{0.359}{0.042}$
& $\textcolor{bestblue}{\mathbf{0.004}}_{\scriptscriptstyle \pm 0.002}$
& $\textcolor{bestblue}{\mathbf{0.004}}_{\scriptscriptstyle \pm 0.002}$ \\

\bottomrule
\end{tabular}%
}

\end{table*}

\section{Conclusions}
We formalize causal fidelity through discrepancies between inferential distributions of a target causal estimand, and show that strong distributional fidelity does not generally guarantee ATE preservation. We introduce a causal-fidelity-aware training framework that augments the generative objective with a causal discrepancy penalty, and instantiate it as a causal-penalized TabDDPM using an on-policy score-function estimator. Across simulation settings and benchmarks, the proposed method improved causal fidelity while largely preserving distributional fidelity. Future work may extend the framework to broader causal estimands, alternative causal inference operators, and other classes of tabular generative models.

\newpage

\section*{AI Use Statement}
The research ideas, methodology, experimental design, and conclusions were developed by the authors. OpenAI Codex was used to assist with code development and deployment, and ChatGPT was used for manuscript editing, language polishing, and refinement of parts of the theoretical analysis. All AI-assisted outputs were reviewed and verified by the authors. The authors take full responsibility for the content, claims, experimental designs, codes, and results in this submission. 

\bibliographystyle{apalike} 
\bibliography{ref} 

\clearpage
\appendix
\thispagestyle{empty}

\section{Experiments Details}
\label{app:exp}

\subsection{Comparison of Baseline Tabular Generative Models}
\label{app:baseline-comparison}

To examine how standard tabular generators differ in their ability to preserve causal information, we compare CTGAN, TVAE, and TabDDPM under the same relatively simple causal data-generating setting. This comparison serves as a preliminary study before considering more challenging data-generating mechanisms and motivates our focus on TabDDPM in the subsequent experiments.

\subsubsection{Experimental Setup}

We use Setting~1 with $d=30$, consisting of 25 continuous and five categorical covariates, randomized treatment $T \sim \mathrm{Bernoulli}(0.5)$, and a constant treatment effect. We consider $\tau \in \{-2,1,5\}$ with Gaussian outcome noise $\sigma=1$. Each reference dataset contains 3,000 observations, split into 2,000 training, 500 validation, and 500 test samples. The data-generating seed is fixed at 42.

For each generator, we perform 10 independent training runs and generate 10 synthetic datasets of 3,000 observations from each fitted model, yielding 100 synthetic datasets per method. Training seeds are varied across fitted models, and distinct sampling seeds are used for each synthetic dataset. Causal fidelity is evaluated using CausalPFN with 8,000 posterior draws, applied consistently to the reference training data and all synthetic datasets.

The final generator configurations are selected under Setting~1 with $\tau=1$. CTGAN uses 300 epochs, a batch size of 500, 128-dimensional embeddings, generator and discriminator dimensions of $(256,256)$, and learning rates of $10^{-4}$. TVAE uses 300 epochs, a batch size of 500, 128-dimensional embeddings, and encoder and decoder dimensions of $(256,256)$; its learning rate is left at the package default. TabDDPM is trained for 50,000 steps with a batch size of 2,000, learning rate $3\times10^{-4}$, two hidden layers of size 256, and 1,000 diffusion steps with a cosine schedule. Numerical variables are quantile-normalized before TabDDPM training. CTGAN and TVAE configurations are selected using validation-set distributional fidelity, while the TabDDPM configuration is selected based on the generative training loss.

\subsubsection{Results}

Table~\ref{tab:setting1-tau1-fidelity} reports conventional distributional-fidelity metrics under Setting~1 with $\tau=1$. TabDDPM achieves the lowest discrepancy across all five reported metrics, while TVAE improves substantially over CTGAN on the numerical covariates and outcome but performs worse on categorical fidelity and correlation structure.

\begin{table}[H]
\centering
\caption{
Distributional fidelity of CTGAN, TVAE, and TabDDPM under Setting~1 with $\tau=1$.
Each entry reports the mean and standard deviation across 100 synthetic datasets
(10 fitted models $\times$ 10 datasets).
}
\label{tab:setting1-tau1-fidelity}

\setlength{\tabcolsep}{4.5pt}
\begin{tabular}{lccccc}
\toprule
Method
& Num. WD $\downarrow$
& Cat. JS $\downarrow$
& Corr. $L_2$ $\downarrow$
& Y $W_1$ $\downarrow$
& T TVD $\downarrow$ \\
\midrule

CTGAN
& $0.067_{\scriptscriptstyle \pm 0.009}$
& $0.043_{\scriptscriptstyle \pm 0.010}$
& $2.030_{\scriptscriptstyle \pm 0.058}$
& $0.824_{\scriptscriptstyle \pm 0.222}$
& $0.043_{\scriptscriptstyle \pm 0.033}$ \\

TVAE
& $0.016_{\scriptscriptstyle \pm 0.001}$
& $0.048_{\scriptscriptstyle \pm 0.007}$
& $2.530_{\scriptscriptstyle \pm 0.176}$
& $0.194_{\scriptscriptstyle \pm 0.094}$
& $0.016_{\scriptscriptstyle \pm 0.013}$ \\

TabDDPM
& $\mathbf{0.007}_{\scriptscriptstyle \pm 0.001}$
& $\mathbf{0.017}_{\scriptscriptstyle \pm 0.004}$
& $\mathbf{1.083}_{\scriptscriptstyle \pm 0.221}$
& $\mathbf{0.104}_{\scriptscriptstyle \pm 0.028}$
& $\mathbf{0.009}_{\scriptscriptstyle \pm 0.007}$ \\

\bottomrule
\end{tabular}
\end{table}

Distributional fidelity alone does not directly assess whether the treatment-effect information relevant to downstream causal inference is preserved. We therefore additionally compare the ATE posteriors inferred from the reference and synthetic datasets.

Figure~\ref{fig:baseline-comparison-1} shows the ATE posteriors under Setting~1 with $\tau=1$. CTGAN exhibits a large shift from the reference posterior and substantial variation across synthetic datasets. TVAE reduces this discrepancy, while TabDDPM produces posteriors that are much more closely aligned with the reference posterior.


Table~\ref{tab:baseline-ate-gap} summarizes the posterior mean discrepancies across the three treatment-effect magnitudes. We report
\[
    \left|
    \mathbb{E}\!\left[\tau \mid D_{\mathrm{syn}}\right]
    -
    \mathbb{E}\!\left[\tau \mid D_{\mathrm{ref}}\right]
    \right|.
\]
TabDDPM yields the smallest discrepancy for all three values of $\tau$, with TVAE showing intermediate performance and CTGAN substantially larger deviations.

\begin{table}[t]
    \centering
    \caption{
   Absolute discrepancy between the synthetic- and reference-derived posterior mean ATEs under Setting~1.
    }
    \label{tab:baseline-ate-gap}
    \begin{tabular}{lccc}
        \toprule
        True $\tau$ & CTGAN & TVAE & TabDDPM \\
        \midrule
        $-2$ & 2.102 & 0.952 & \textbf{0.109} \\
        $1$  & 0.896 & 0.328 & \textbf{0.085} \\
        $5$  & 4.806 & 0.715 & \textbf{0.199} \\
        \bottomrule
    \end{tabular}
\end{table}

Overall, TabDDPM provides the strongest baseline in this setting in terms of both conventional distributional fidelity and preservation of the ATE posterior. We therefore focus on TabDDPM in the subsequent experiments and examine whether this behavior remains reliable under more challenging data-generating mechanisms.

\subsection{Causal Fidelity of Vanilla TabDDPM}
\label{append: vanilla_tabddpm}

We next examine whether the strong baseline performance of vanilla TabDDPM remains consistent across a broader range of data-generating regimes and dimensions.

\subsubsection{Experimental Setup}

We use the same three families of simulation mechanisms described in Appendix~\ref{append:dgp}, covering linear outcomes, nonlinear outcomes, and heterogeneous treatment effects. Here, we vary the covariate dimension over $d\in\{10,20,30\}$ to examine the stability of vanilla TabDDPM across problem sizes and data-generating regimes.

The main experiments use mixed continuous and categorical covariates. The $d=10$, $d=20$, and $d=30$ settings contain 8 continuous and 2 categorical variables, 16 continuous and 4 categorical variables, and 25 continuous and 5 categorical variables, respectively. Treatment remains randomized as $T\sim\mathrm{Bernoulli}(0.5)$, and the outcome noise level is fixed at $\sigma=1$. For each configuration, we train 10 independent TabDDPM models and generate 10 synthetic datasets from each fitted model.

The TabDDPM configuration varies with dimension. For $d=30$, the model is trained for 50,000 steps with a batch size of 2,000, learning rate $3\times10^{-4}$, hidden-layer dimensions $(256,256)$, and 1,000 diffusion steps. For $d=20$, we use 30,000 training steps with hidden-layer dimensions $(512,256)$; for $d=10$, we use 30,000 steps with hidden-layer dimensions $(256,256)$. Both retain the same batch size, learning rate, and number of diffusion steps as the $d=30$ configuration. Each model is trained on 2,000 observations and generates synthetic datasets of size 3,000.

Causal fidelity is evaluated using CausalPFN-derived ATE posteriors. For $d=30$, we use direct ATE inference with 8,000 posterior draws. For $d=10$ and $d=20$, we use an 80/20 split for CATE-based inference and aggregate the estimated CATEs to obtain the ATE, using 2,000 posterior draws. Figure~\ref{fig:tabddpm-representative} additionally includes one $d=10$ all-continuous example generated under the corresponding all-continuous configuration.

\subsubsection{Results}

\begin{figure}[t]
    \centering
    \includegraphics[width=0.78\textwidth]{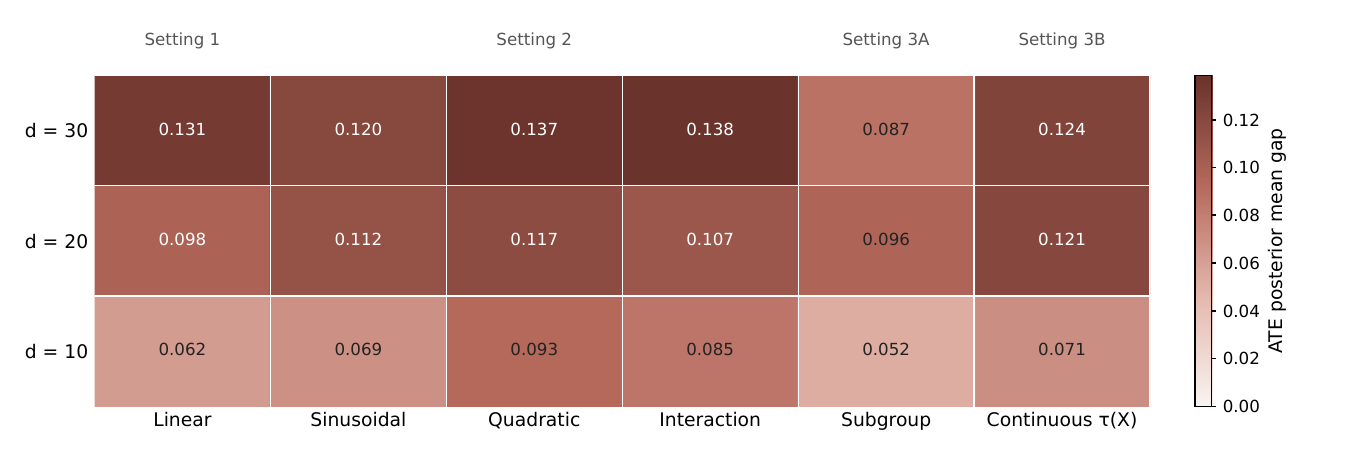}
    \caption{Causal-fidelity discrepancy of vanilla TabDDPM across simulation regimes and dimensions. Each cell shows the mean absolute synthetic--reference ATE posterior mean gap, averaged over treatment-effect configurations.}
    \label{fig:tabddpm-heatmap}
\end{figure}

Figure~\ref{fig:tabddpm-heatmap} summarizes the causal-fidelity discrepancy across the mixed-covariate simulation regimes. Each cell reports
\[
    \frac{1}{K}
    \sum_{k=1}^{K}
    \left|
    \mathbb{E}\!\left[\tau \mid D_{\mathrm{syn}}^{(k)}\right]
    -
    \mathbb{E}\!\left[\tau \mid D_{\mathrm{ref}}^{(k)}\right]
    \right|,
\]
where $K$ denotes the number of treatment-effect configurations evaluated within the corresponding regime. Setting~3A contains a single subgroup-treatment-effect configuration and is therefore represented by its corresponding discrepancy.

The discrepancy varies across both dimensions and data-generating mechanisms. Larger gaps occur in several higher-dimensional settings, but the pattern is not uniformly monotonic in either dimension or model complexity.

\begin{figure}[t]
    \centering
    \includegraphics[width=\textwidth]{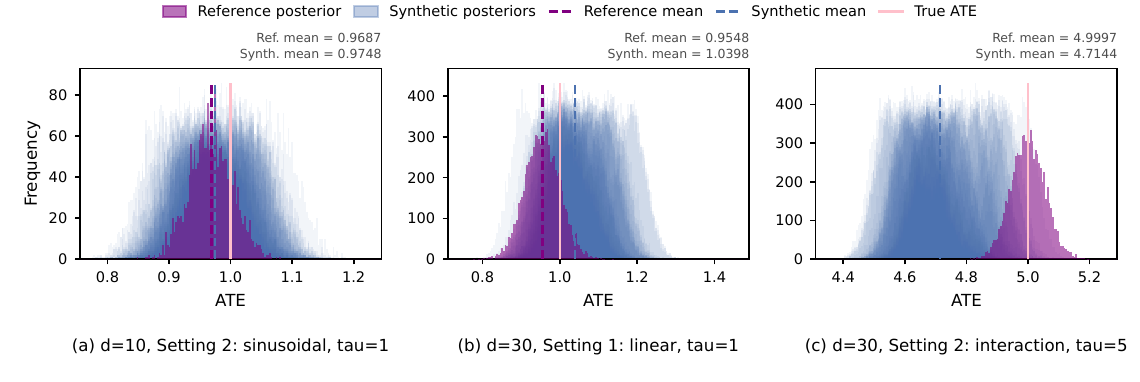}
    \caption{
    Representative ATE posterior comparisons for vanilla TabDDPM. Red and blue denote reference and synthetic posteriors; dashed lines indicate the corresponding posterior means and the true ATE.
    }
    \label{fig:tabddpm-representative}
\end{figure}

Figure~\ref{fig:tabddpm-representative} illustrates this variation with three representative cases. In the $d=10$ sinusoidal setting with $\tau=1$, the synthetic and reference ATE posteriors are closely aligned. In the $d=30$ linear setting with $\tau=1$, the synthetic posteriors remain close to the reference but show a visible shift toward larger ATE values. The $d=30$ interaction setting with $\tau=5$ exhibits a substantially larger and more systematic displacement from the reference posterior.

Overall, vanilla TabDDPM preserves causal information well in some settings but does not do so uniformly across data-generating regimes. This variability motivates incorporating causal fidelity directly into the generative objective rather than relying on distributional fidelity alone.

\subsection{Simulation Study}
\label{append: simulation}

\subsubsection{Simulation Data-Generating Mechanisms}
\label{append:dgp}

Across all settings, the outcome is generated from continuous covariates
$X=(X_1,\ldots,X_{25})^\top$, five categorical covariates
$C_1,\ldots,C_5$, a randomized binary treatment $T$, and Gaussian noise
$\epsilon$. For Settings 1 and 2, the outcome model is
\[
Y
=
f(X)
+
\sum_{j=1}^{5}\alpha_{j,C_j}
+
\tau T
+
\epsilon,
\]
where $\alpha_{j,C_j}$ denotes the level-specific effect of the $j$th
categorical covariate.

\paragraph{Setting 1: Linear outcome model.}
The baseline outcome function is linear,
\[
f_{\mathrm{lin}}(X)=\beta^\top X,
\]
with
\[
\beta
=
(1,\;0.5,\;-0.5,\;0.3,\;-0.3,
\underbrace{0.2,\ldots,0.2}_{20})^\top.
\]

\paragraph{Setting 2: Nonlinear outcome models.}
We consider two nonlinear baseline outcome functions:
\begin{align*}
    f_{\mathrm{sin}}(X) &=\sin(X_1)+0.5X_2+0.3X_3,\\
    f_{\mathrm{quad}}(X) &= \beta^\top X+0.5X_1^2.
\end{align*}
\paragraph{Setting 3: Heterogeneous treatment effects.}
To introduce subgroup-specific treatment-effect heterogeneity, we replace
the first observed categorical covariate by
\[
G \sim \mathrm{Bernoulli}(0.5),
\]
and define the treatment effect as
\[
\tau(G)=
\begin{cases}
3, & G=1,\\
0.1, & G=0.
\end{cases}
\]
The outcome is generated as
\[
Y
=
f_{\mathrm{sin}}(X)
+
\alpha_{1,\widetilde C_1}
+
\sum_{j=2}^{5}\alpha_{j,C_j}
+
\tau(G)T
+
\epsilon,
\]
where
\[
f_{\mathrm{sin}}(X)
=
\sin(X_1)+0.5X_2+0.3X_3.
\]
Here, $\widetilde C_1$ is an auxiliary three-level categorical variable
used only to provide an additive categorical contribution to the outcome.
Each category coefficient is drawn from
$\mathcal N(0,0.5^2)$ and then held fixed across observations. Thus,
$G$ controls treatment-effect heterogeneity, while
$\widetilde C_1$ contributes additional outcome variation. Since
$G\sim\mathrm{Bernoulli}(0.5)$, the population ATE is
\[
\mathbb E[\tau(G)]
=
0.5\times 3 + 0.5\times 0.1
=
1.55.
\]

\subsubsection{Implementation Details}

Both vanilla and causal-penalized TabDDPM use Gaussian diffusion for numerical variables and multinomial diffusion for categorical variables, with $1,000$ diffusion steps. The denoising network uses a $128$-dimensional time embedding and two hidden layers of $256$ units with ReLU activations and no dropout. Models are trained for $20,000$ updates using AdamW with batch size $2,000$, an initial learning rate of $3\times10^{-4}$ with linear decay, and zero weight decay. 

For causal-penalized TabDDPM, the causal discrepancy is the Wasserstein-1 distance between the CausalPFN ATE posteriors inferred from the reference and synthetic datasets. After 10{,}000 warm-up updates, causal updates are performed every 100 iterations using $m=10$ synthetic tables of 2{,}000 observations, penalty weight $\lambda=1$, and moving-baseline rate $\rho=0.2$. 

For each method and simulation scenario, we fit one model using training seed 0 and evaluate the final model on 20 synthetic datasets of 3{,}000 observations each, using matched sampling seeds across methods.

\subsubsection{Additional Results}
To further assess robustness, we compare vanilla TabDDPM and Causal-TabDDPM across multiple causal outcome models and treatment-effect values. As shown in Table~\ref{tab:additional_causal_results}, Causal-TabDDPM generally achieves lower MAE, RMSE, and posterior Wasserstein-1 distance across most of the considered settings, indicating more accurate preservation of the target ATE. The gains are particularly pronounced in settings where vanilla TabDDPM exhibits larger treatment-effect discrepancies.

\begin{table*}[t]
\centering
\caption{
Causal fidelity comparison of vanilla TabDDPM and Causal-TabDDPM across multiple simulation settings and treatment-effect values. The better result within each setting is shown in bold.
}
\label{tab:additional_causal_results}

\scriptsize
\setlength{\tabcolsep}{4pt}

\resizebox{\textwidth}{!}{%
\begin{tabular}{@{}llccccc@{}}
\toprule
\textbf{Setting}
& \textbf{True ATE}
& \textbf{Method}
& \textbf{ATE mean $\pm$ SD}
& \textbf{MAE $\downarrow$}
& \textbf{RMSE $\downarrow$}
& \textbf{Post. W$_1$ $\downarrow$} \\
\midrule

\multirow{2}{*}{Linear}
& \multirow{2}{*}{$-2$}
& TabDDPM
& $-1.7099 \pm 0.0855$
& $0.2901$
& $0.3018$
& $0.3361$ \\
&
& Causal-TabDDPM
& $\mathbf{-2.1165} \pm 0.1388$
& $\mathbf{0.1454}$
& $\mathbf{0.1786}$
& $\mathbf{0.1233}$ \\
\midrule

\multirow{2}{*}{Linear}
& \multirow{2}{*}{$1$}
& TabDDPM
& $\mathbf{0.9729} \pm 0.0614$
& $\mathbf{0.0499}$
& $\mathbf{0.0657}$
& $\mathbf{0.0532}$ \\
&
& Causal-TabDDPM
& $0.9517 \pm 0.0618$
& $0.0609$
& $0.0772$
& $0.0546$ \\
\midrule

\multirow{2}{*}{Sin mix}
& \multirow{2}{*}{$1$}
& TabDDPM
& $1.2005 \pm 0.0738$
& $0.2005$
& $0.2130$
& $0.2370$ \\
&
& Causal-TabDDPM
& $\mathbf{0.9728} \pm 0.0680$
& $\mathbf{0.0578}$
& $\mathbf{0.0716}$
& $\mathbf{0.0575}$ \\
\midrule

\multirow{2}{*}{Sin mix}
& \multirow{2}{*}{$5$}
& TabDDPM
& $4.7801 \pm 0.1043$
& $\mathbf{0.2233}$
& $\mathbf{0.2422}$
& $\mathbf{0.1930}$ \\
&
& Causal-TabDDPM
& $\mathbf{4.8066} \pm 0.2010$
& $0.2474$
& $0.2753$
& $0.2296$ \\
\midrule

\multirow{2}{*}{Quadratic}
& \multirow{2}{*}{$1$}
& TabDDPM
& $1.2084 \pm 0.1077$
& $0.2084$
& $0.2334$
& $0.2127$ \\
&
& Causal-TabDDPM
& $\mathbf{1.0108} \pm 0.0877$
& $\mathbf{0.0600}$
& $\mathbf{0.0862}$
& $\mathbf{0.0655}$ \\
\midrule

\multirow{2}{*}{Quadratic}
& \multirow{2}{*}{$5$}
& TabDDPM
& $5.1354 \pm 0.1117$
& $0.1443$
& $\mathbf{0.1738}$
& $0.1487$ \\
&
& Causal-TabDDPM
& $\mathbf{5.0382} \pm 0.1760$
& $\mathbf{0.1350}$
& $0.1758$
& $\mathbf{0.1413}$ \\
\midrule

\multirow{2}{*}{Heterogeneous}
& \multirow{2}{*}{$1.55$}
& TabDDPM
& $1.6677 \pm 0.0693$
& $0.1181$
& $0.1357$
& $0.1247$ \\
&
& Causal-TabDDPM
& $\mathbf{1.4978} \pm 0.0635$
& $\mathbf{0.0641}$
& $\mathbf{0.0810}$
& $\mathbf{0.0608}$ \\

\bottomrule
\end{tabular}%
}

\end{table*}

\subsection{Real Application}
\label{append: application}

\subsubsection{Implementation Details}

For both IHDP and IST, each generator is fitted once and used to produce 20 synthetic datasets, each containing 3,000 samples. We use a fixed training seed of 0 and sampling seeds from 10,000 to 10,019. The same train--validation--test split is used across methods, with a split seed of 43. Causal fidelity is evaluated using CausalPFN with 8,000 posterior draws and a fixed inference seed of 314159. For all methods, the reference ATE posterior is estimated from the training split, and the same inference procedure is applied to every synthetic dataset.

CTGAN and TVAE are implemented using SDV 1.18.0, while TabSyn follows the official ICLR 2024 implementation. Their hyperparameters are selected separately on IHDP and IST using validation-set distributional fidelity, without using the causal-fidelity metrics for model selection. The final configurations are summarized in Table~\ref{tab:realapp-config}. For CTGAN and TVAE, the main tuning dimensions include network size, batch size, learning rate or latent dimension, and training length. For TabSyn, we tune the VAE regularization strength, batch size, and diffusion model width. Final TabSyn models are trained for 4,000 VAE epochs and 10,001 diffusion epochs, with 50 sampling steps.

\begin{table}[h]
\centering
\caption{
Final generator configurations used in the real-application experiments.
Only the main tuned parameters are shown.
}
\label{tab:realapp-config}
\small
\setlength{\tabcolsep}{4.5pt}
\begin{tabular}{lll}
\toprule
Method & Dataset & Key parameters \\
\midrule
CTGAN
& IHDP
& lr=$10^{-4}$, batch=500, epochs=300 \\

CTGAN
& IST
& lr=$10^{-4}$, batch=200, epochs=300 \\

TVAE
& IHDP
& dims=(128,128), batch=200, epochs=300 \\

TVAE
& IST
& dims=(256,256), batch=500, epochs=300 \\

TabSyn
& IHDP
& $\beta_{\max}=0.05$, batch=4096 \\

TabSyn
& IST
& $\beta_{\max}=0.01$, batch=1024 \\

TabDDPM
& IHDP
& layers=[512,1024,1024,1024,1024,512], diffusion steps=1000 \\

TabDDPM
& IST
& layers=[256,128], diffusion steps=100 \\
\bottomrule
\end{tabular}
\end{table}

Vanilla TabDDPM and Causal-TabDDPM use the same generator architecture, diffusion settings, optimization schedule, training budget, initialization seed, and evaluation protocol within each dataset. The IHDP models are trained for 20,000 steps with a batch size of 498, 1,000 diffusion steps, and a learning rate of $10^{-3}$; the corresponding IST settings use a batch size of 4,096 and 100 diffusion steps. Causal-TabDDPM adds on-policy causal regularization with $\lambda=1$, using $m=10$ synthetic trajectories per causal update and a moving-average baseline with $\rho=0.2$. Causal updates begin after 10,000 training steps and are applied every 100 steps. The discrepancy is the Wasserstein-1 distance between the ATE posteriors inferred from the reference and synthetic datasets. The causal gradient is capped relative to the generative gradient, with maximum ratios of 1.1 for IHDP and 1.0 for IST.

\subsubsection{Additional Results}
\label{append:application-results}
 
\begin{figure*}[t]
    \centering
    \includegraphics[width=\textwidth]{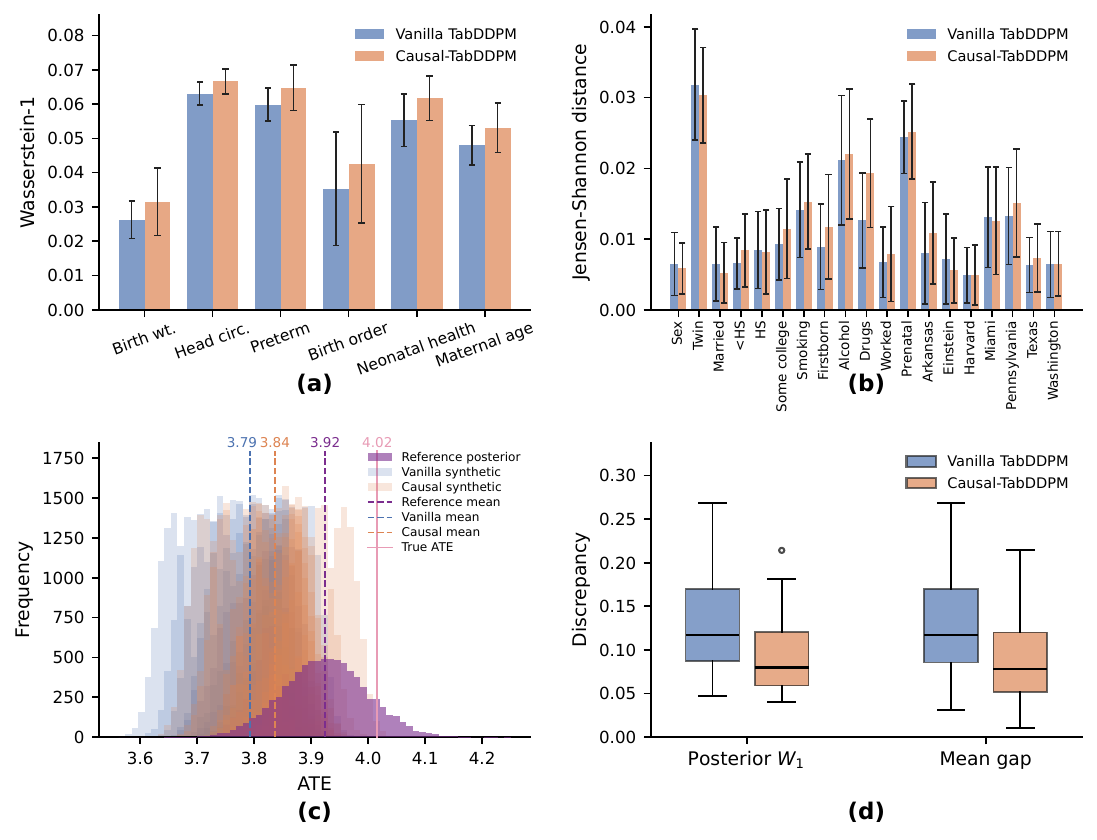}
    \caption{
    Detailed comparison of vanilla TabDDPM and Causal-TabDDPM on IHDP. Panels (a)-(b) show feature-wise distributional discrepancies, panel (c) compares ATE posteriors, and panel (d) reports posterior Wasserstein distance and absolute posterior mean gap across synthetic datasets.
    }
    \label{fig:ihdp-additional}
\end{figure*}

\begin{figure*}[t]
    \centering
    \includegraphics[width=\textwidth]{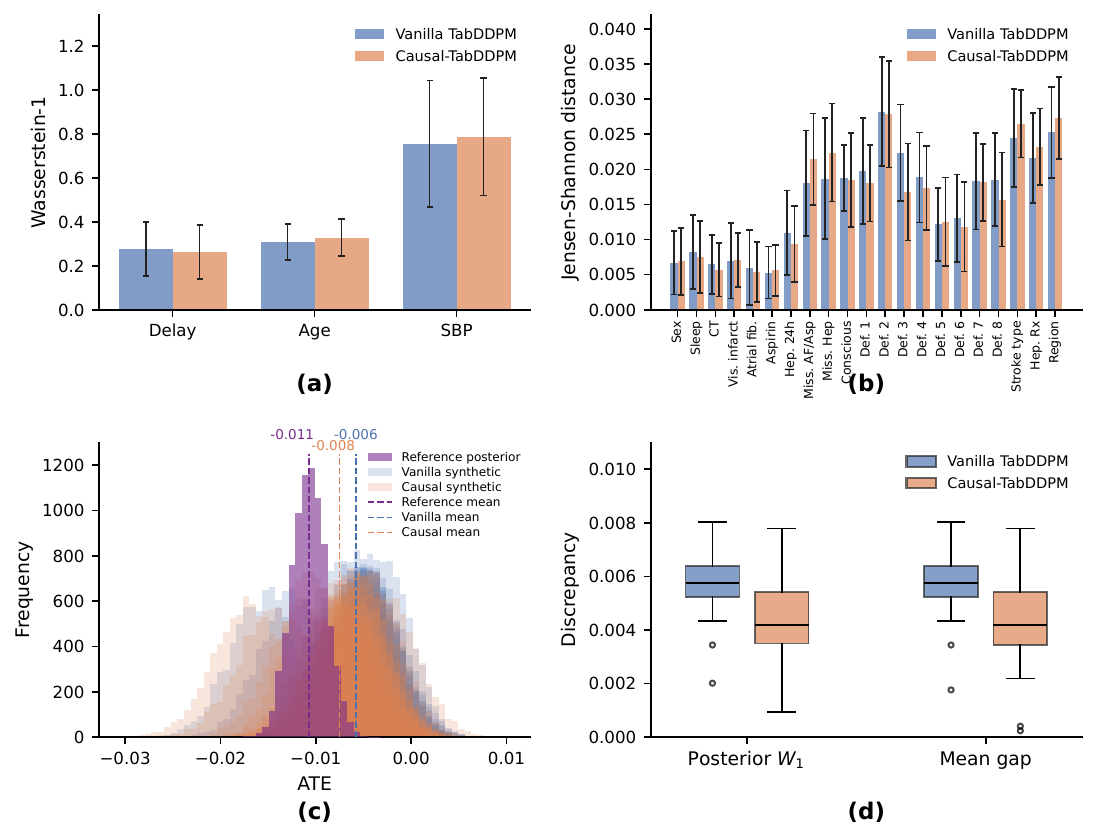}
    \caption{
    Detailed comparison of vanilla TabDDPM and Causal-TabDDPM on IST. Panels (a)-(b) show feature-wise distributional discrepancies, panel (c) compares ATE posteriors, and panel (d) reports posterior Wasserstein distance and absolute posterior mean gap across synthetic datasets.
    }
    \label{fig:ist-additional}
\end{figure*}

Figures~\ref{fig:ihdp-additional} and~\ref{fig:ist-additional} provide a more detailed comparison between vanilla TabDDPM and Causal-TabDDPM. On IHDP, causal regularization generally leads to slightly larger feature-level discrepancies, particularly for the numerical covariates. The differences are small, however, and are not uniform across all categorical variables. In contrast, the causal metrics improve more consistently: the synthetic ATE posteriors move toward the reference posterior, and both the posterior Wasserstein distance and posterior mean gap decrease.

The pattern on IST is somewhat different. Feature-level distributional fidelity remains close between the two models, with some covariates improving under causal regularization and others showing small increases in discrepancy. This suggests that the effect on distributional fidelity is not uniformly negative. The causal improvement is again more systematic. The synthetic ATE posteriors shift toward the reference posterior, and the distributions of both causal discrepancy measures move downward relative to vanilla TabDDPM. These visualizations complement the aggregate results in Table~\ref{tab:real_application} by showing that the gains in causal fidelity do not arise from a uniform deterioration in marginal data fidelity.

\subsection{Computing Infrastructure}
\label{app:computing}
All experiments were conducted on a workstation equipped with an NVIDIA GeForce RTX 4090 GPU. Model training, synthetic-data generation, and causal fidelity evaluation were performed using this GPU environment.

\clearpage
\begingroup
\section{Proofs of the Theoretical Results}
\label{app:theory-proofs}

\subsection{Proof of Theorem~\ref{thm:joint-ate-gap}}
\label{app:theory-counterexample}

\textbf{Identification and ATEs.}
Fix $\epsilon\in(0,1)$. Under $\mathcal P$, choose potential
outcomes $Y(0)=Y(1)=U$; under $\widehat{\mathcal P}_\theta$ choose
$\widetilde Y(0)=\widetilde U$ and
$\widetilde Y(1)=\widetilde U+\Delta$.
Consistency holds in both cases. Each pair of potential outcomes is
independent of its treatment indicator conditional on its covariate.
Moreover, $\mathbb P(W=1\mid X)=\epsilon\in(0,1)$ under
$\mathcal P$, and
$\mathbb P(\widetilde W=1\mid\widetilde X)=\epsilon$
under $\widehat{\mathcal P}_\theta$.
The identification assumptions therefore hold, with $\tau=0$ and
$\widetilde\tau_\theta=\Delta$.

\textbf{Total variation and KL divergence.}
The joint distributions of $(X,W)$ and $(\widetilde X,\widetilde W)$
coincide, as do the conditional outcome distributions in the control arm.
In the treatment arm, the real and generated outcome laws are
$\mathcal N(0,\sigma^2)$ and $\mathcal N(\Delta,\sigma^2)$,
respectively. Let $\phi_\sigma$ be the density of
$\mathcal N(0,\sigma^2)$ and $\Phi$ the standard normal distribution
function. Integrating out the common covariate density and summing
over treatment status yields
\begin{align*}
d_{\mathrm{TV}}(\mathcal P,\widehat{\mathcal P}_\theta)
&=\frac{\epsilon}{2}\int_{\mathbb R}
\bigl|\phi_\sigma(y)-\phi_\sigma(y-\Delta)\bigr|\,dy\\
&=\epsilon\left[2\Phi\!\left(\frac{\Delta}{2\sigma}\right)-1\right]
\leq\epsilon.
\end{align*}
The second equality follows because the Gaussian densities cross at
$y=\Delta/2$. For KL divergence, the common covariate and treatment
factors cancel in the density ratio, giving
\begin{align*}
D_{\mathrm{KL}}(\mathcal P\Vert\widehat{\mathcal P}_\theta)
&=\epsilon\,\mathbb E_{Y\sim\mathcal N(0,\sigma^2)}
\left[\frac{(Y-\Delta)^2-Y^2}{2\sigma^2}\right]
=\frac{\epsilon\Delta^2}{2\sigma^2},\\
D_{\mathrm{KL}}(\widehat{\mathcal P}_\theta\Vert\mathcal P)
&=\epsilon\,\mathbb E_{\widetilde Y\sim\mathcal N(\Delta,\sigma^2)}
\left[\frac{\widetilde Y^2-(\widetilde Y-\Delta)^2}{2\sigma^2}\right]
=\frac{\epsilon\Delta^2}{2\sigma^2}.
\end{align*}

\textbf{Wasserstein distance.}
Couple the laws by setting $\widetilde X=X$, $\widetilde W=W$,
and $\widetilde U=U$. The resulting Euclidean transport cost is
$|\Delta W|$, so
$W_1(\mathcal P,\widehat{\mathcal P}_\theta)\leq\epsilon\Delta$.
Conversely, every coupling of
$(X,W,Y)\sim\mathcal P$ and
$(\widetilde X,\widetilde W,\widetilde Y)
\sim\widehat{\mathcal P}_\theta$ satisfies
\[
\mathbb E\bigl\|(X,W,Y)-(\widetilde X,\widetilde W,\widetilde Y)\bigr\|_2
\geq\mathbb E|Y-\widetilde Y|
\geq|\mathbb EY-\mathbb E\widetilde Y|
=\epsilon\Delta.
\]
Taking the infimum over couplings completes the
proof.\hfill$\square$

\textbf{Connection to inferred ATE distributions.}
Fix $\epsilon\in(0,1)$ and consider
$D_{\mathrm{real}}\sim\mathcal P^{\,n}$ and
$D_{\mathrm{syn}}\sim\widehat{\mathcal P}_\theta^{\,n_{\mathrm{syn}}}$.
Recall that
$\widehat p_{\mathrm{real}}=\mathcal A(D_{\mathrm{real}})$ and
$\widehat p_{\mathrm{syn}}=\mathcal A(D_{\mathrm{syn}})$.
Denote a point mass at $a$ by $\delta_a$.
The triangle inequality and its reverse form imply
\[
\left|W_1(\widehat p_{\mathrm{real}},\widehat p_{\mathrm{syn}})
-|\widetilde\tau_\theta-\tau|\right|
\leq W_1(\widehat p_{\mathrm{real}},\delta_\tau)
+W_1(\widehat p_{\mathrm{syn}},\delta_{\widetilde\tau_\theta}),
\]
because $W_1(\delta_\tau,\delta_{\widetilde\tau_\theta})
=|\widetilde\tau_\theta-\tau|=\Delta$.
If both terms on the right converge to zero in probability as
$n,n_{\mathrm{syn}}\to\infty$, then
$\mathcal D_\tau(D_{\mathrm{real}},D_{\mathrm{syn}})$ with $d=W_1$
converges in probability to $\Delta$.
The order of limits is essential. For equal, fixed sample sizes
$n_{\mathrm{syn}}=n$,
\[
d_{\mathrm{TV}}(\mathcal P^{\,n},
\widehat{\mathcal P}_\theta^{\,n})
\leq n\,d_{\mathrm{TV}}(\mathcal P,\widehat{\mathcal P}_\theta)
\leq n\epsilon.
\]
Consequently, taking $\epsilon\downarrow0$ at fixed sample size
does not establish separation of the inferred ATE distributions.

\subsection{Proof of Theorem~\ref{thm:causal-penalty}}
\label{app:theory-regularization}

\textbf{Comparator bound and strict improvement.}
By Equation~\eqref{eq:theory-approx-opt}, any
$\bar\theta\in\Theta$ satisfies
\[
\mathcal L_{\mathrm{gen}}(\theta_\lambda)
+\lambda\mathcal L_{\mathrm{causal}}(\theta_\lambda)
\leq\mathcal L_{\mathrm{gen}}(\bar\theta)
+\lambda\mathcal L_{\mathrm{causal}}(\bar\theta)
+\delta_{\mathrm{opt}}.
\]
Rearranging and using
$\mathcal L_{\mathrm{gen}}(\theta_\lambda)
\geq\mathcal L_{\mathrm{gen}}^\star$ gives
\begin{align*}
\lambda\bigl\{\mathcal L_{\mathrm{causal}}(\theta_\lambda)
-\mathcal L_{\mathrm{causal}}(\bar\theta)\bigr\}
&\leq\mathcal L_{\mathrm{gen}}(\bar\theta)
-\mathcal L_{\mathrm{gen}}(\theta_\lambda)+\delta_{\mathrm{opt}}\\
&\leq\mathcal L_{\mathrm{gen}}(\bar\theta)
-\mathcal L_{\mathrm{gen}}^\star+\delta_{\mathrm{opt}}.
\end{align*}
Division by $\lambda>0$ proves Equation~\eqref{eq:theory-oracle}.
If $\theta_{\mathrm{gen}}$ minimizes $\mathcal L_{\mathrm{gen}}$,
substituting $\bar\theta=\theta_{\mathrm{gen}}$ gives
\[
\mathcal L_{\mathrm{causal}}(\theta_\lambda)
\leq\mathcal L_{\mathrm{causal}}(\theta_{\mathrm{gen}})
+\delta_{\mathrm{opt}}/\lambda.
\]
For a comparator with
$\mathcal L_{\mathrm{causal}}(\bar\theta)
<\mathcal L_{\mathrm{causal}}(\theta_{\mathrm{gen}})$,
Equation~\eqref{eq:theory-strict-improvement} makes the right-hand
side of Equation~\eqref{eq:theory-oracle} strictly smaller than
$\mathcal L_{\mathrm{causal}}(\theta_{\mathrm{gen}})$,
proving strict improvement.

\textbf{Monotonicity and the generative tradeoff.}
For $0\leq\lambda_1<\lambda_2$, exact optimality at each weight gives
\begin{align*}
\mathcal L_{\mathrm{gen}}(\theta_{\lambda_1})
+\lambda_1\mathcal L_{\mathrm{causal}}(\theta_{\lambda_1})
&\leq\mathcal L_{\mathrm{gen}}(\theta_{\lambda_2})
+\lambda_1\mathcal L_{\mathrm{causal}}(\theta_{\lambda_2}),\\
\mathcal L_{\mathrm{gen}}(\theta_{\lambda_2})
+\lambda_2\mathcal L_{\mathrm{causal}}(\theta_{\lambda_2})
&\leq\mathcal L_{\mathrm{gen}}(\theta_{\lambda_1})
+\lambda_2\mathcal L_{\mathrm{causal}}(\theta_{\lambda_1}).
\end{align*}
Adding these inequalities yields
\[
(\lambda_2-\lambda_1)\bigl\{
\mathcal L_{\mathrm{causal}}(\theta_{\lambda_2})
-\mathcal L_{\mathrm{causal}}(\theta_{\lambda_1})\bigr\}\leq0.
\]
Since $\lambda_2>\lambda_1$, this proves
Equation~\eqref{eq:theory-causal-monotone}. The same optimality
inequalities imply
\begin{align*}
0&\leq\lambda_1\bigl\{
\mathcal L_{\mathrm{causal}}(\theta_{\lambda_1})
-\mathcal L_{\mathrm{causal}}(\theta_{\lambda_2})\bigr\}\\
&\leq\mathcal L_{\mathrm{gen}}(\theta_{\lambda_2})
-\mathcal L_{\mathrm{gen}}(\theta_{\lambda_1})\\
&\leq\lambda_2\bigl\{
\mathcal L_{\mathrm{causal}}(\theta_{\lambda_1})
-\mathcal L_{\mathrm{causal}}(\theta_{\lambda_2})\bigr\}.
\end{align*}
This also proves Equation~\eqref{eq:theory-generative-monotone}.
In particular, comparing an exact penalized optimum with
$\theta_{\mathrm{gen}}$ yields
\[
0\leq\mathcal L_{\mathrm{gen}}(\theta_\lambda)
-\mathcal L_{\mathrm{gen}}(\theta_{\mathrm{gen}})
\leq\lambda\bigl\{
\mathcal L_{\mathrm{causal}}(\theta_{\mathrm{gen}})
-\mathcal L_{\mathrm{causal}}(\theta_\lambda)\bigr\}.
\]
These inequalities require neither convexity nor correct
specification of the generator class.\hfill$\square$

\subsection{From Causal Discrepancy to ATE Error}
\label{app:theory-ate-transfer}

For any two probability laws $\mu$ and $\nu$ on the real line
with finite first moments, every coupling $\pi$ satisfies
\[
\left|\int u\,\mu(du)-\int v\,\nu(dv)\right|
=\left|\int(u-v)\,\pi(du,dv)\right|
\leq\int|u-v|\,\pi(du,dv).
\]
Taking the infimum over couplings shows that $W_1(\mu,\nu)$
bounds the absolute difference between their means. Applying this
inequality to the inferred ATE distributions gives
\[
|\widehat\tau_{\mathrm{syn}}-\widehat\tau_{\mathrm{real}}|
\leq W_1(\widehat p_{\mathrm{syn}},\widehat p_{\mathrm{real}})
=\mathcal D_\tau(D_{\mathrm{real}},D_{\mathrm{syn}}).
\]
The triangle inequality followed by expectation over
$D_{\mathrm{syn}}\sim\widehat{\mathcal P}_\theta^{\,n_{\mathrm{syn}}}$
therefore proves Equation~\eqref{eq:theory-mean-error}, with
$e_{\mathrm{real}}=|\widehat\tau_{\mathrm{real}}-\tau|$.
For a penalized solution and any $\bar\theta\in\Theta$,
Theorem~\ref{thm:causal-penalty} gives
\begin{align*}
\mathbb E_{D_{\mathrm{syn}}\sim
\widehat{\mathcal P}_{\theta_\lambda}^{\,n_{\mathrm{syn}}}}
\bigl|\widehat\tau_{\mathrm{syn}}-\tau\bigr|
\leq e_{\mathrm{real}}
+\mathcal L_{\mathrm{causal}}(\bar\theta)+\frac{\mathcal L_{\mathrm{gen}}(\bar\theta)
-\mathcal L_{\mathrm{gen}}^\star+\delta_{\mathrm{opt}}}{\lambda}.
\end{align*}

To bound the population ATE discrepancy, suppose the generated law
$\widehat{\mathcal P}_\theta$ satisfies the identification conditions
and has finite population ATE $\widetilde\tau_\theta$.
For the fixed synthetic sample size $n_{\mathrm{syn}}$, define the
expected absolute estimation error under this law by
\[
e_{\mathrm{syn}}(\theta)
:=\mathbb E_{D_{\mathrm{syn}}\sim
\widehat{\mathcal P}_\theta^{\,n_{\mathrm{syn}}}}
\bigl|\widehat\tau_{\mathrm{syn}}-\widetilde\tau_\theta\bigr|.
\]
The triangle inequality and expectation over $D_{\mathrm{syn}}$ yield
\begin{equation}
|\widetilde\tau_\theta-\tau|
\leq e_{\mathrm{syn}}(\theta)
+\mathcal L_{\mathrm{causal}}(\theta)+e_{\mathrm{real}}.
\label{eq:theory-population-transfer}
\end{equation}
Equation~\eqref{eq:theory-population-transfer} bounds the population
ATE discrepancy by three terms: synthetic estimation error, causal
discrepancy, and reference estimation error.
In particular, if
$e_{\mathrm{syn}}(\theta)\leq\bar e_{\mathrm{syn}}$ uniformly over
the compared generators, then
$\bar e_{\mathrm{syn}}+e_{\mathrm{real}}
+\mathcal L_{\mathrm{causal}}(\theta)$ is a common valid upper bound.
Along exact penalized optima, this common upper bound is nonincreasing
in $\lambda$. This conclusion concerns the error bound; the actual
population ATE discrepancy need not be monotone in $\lambda$.

\subsection{Proof of the Gaussian Example}
\label{app:theory-gaussian-example}

Under $\mathcal P$, take $Y(0)=\beta_0+U$ and
$Y(1)=\beta_0+\tau+U$; under $\widehat{\mathcal P}_\theta$,
take $\widetilde Y(0)=\widetilde U$ and
$\widetilde Y(1)=\theta+\widetilde U$.
Randomized treatment with probability $1/2$ ensures identification,
so the respective ATEs are $\tau$ and $\widetilde\tau_\theta=\theta$.
The covariate and treatment distributions coincide under the two laws.
The KL divergence between normal distributions with common variance
$\sigma^2$ is their squared mean difference divided by $2\sigma^2$.
Conditioning on treatment therefore gives
\[
\mathcal L_{\mathrm{gen}}(\theta)
=\frac{\beta_0^2}{4\sigma^2}
+\frac{(\theta-\tau-\beta_0)^2}{4\sigma^2},
\qquad \theta_{\mathrm{gen}}=\tau+\beta_0.
\]
The generator's fixed zero intercept forces the unpenalized fit to
absorb $\beta_0$ into its treatment coefficient.

Since $W_1(\delta_\tau,\delta_\theta)=|\theta-\tau|$, setting
$v=\theta-\tau$ reduces the population regularized objective,
up to an additive constant, to
\[
\frac{(v-\beta_0)^2}{4\sigma^2}+\lambda|v|.
\]
This function is strictly convex. Its unique minimizer satisfies
the subgradient condition
\[
0\in\frac{v-\beta_0}{2\sigma^2}+\lambda\,\partial|v|.
\]
For $v\ne0$, solving this condition gives
$v=\beta_0-2\lambda\sigma^2\operatorname{sgn}(v)$;
the solution is $v=0$ exactly when
$|\beta_0|\leq2\lambda\sigma^2$.
Together these cases yield Equation~\eqref{eq:theory-gaussian-solution}
and the stated ATE error. Since $\beta_0\ne0$, this error is strictly
smaller than $|\beta_0|$ for every $\lambda>0$.
The corresponding increase in generative loss is also explicit:
\[
\mathcal L_{\mathrm{gen}}(\theta_\lambda)
-\mathcal L_{\mathrm{gen}}(\theta_{\mathrm{gen}})
=\frac{\min\{|\beta_0|,\,2\lambda\sigma^2\}^{2}}{4\sigma^2}.
\]
Thus causal fidelity improves through a quantifiable tradeoff in
joint distributional fit.\hfill$\square$

\textbf{Relation to the finite-sample penalty.}
The example uses a population analogue of the Wasserstein penalty.
For each fixed $\theta$, the reverse triangle inequality and
expectation over $D_{\mathrm{syn}}$ give
\[
\left|\mathcal L_{\mathrm{causal}}(\theta)-|\theta-\tau|\right|
\leq W_1(\widehat p_{\mathrm{real}},\delta_\tau)
+\mathbb E_{D_{\mathrm{syn}}\sim
\widehat{\mathcal P}_\theta^{\,n_{\mathrm{syn}}}}
W_1(\widehat p_{\mathrm{syn}},\delta_\theta).
\]
The population penalty is recovered when both terms on the right
vanish, requiring convergence in mean on the synthetic side.
The closed-form solution therefore describes the population
illustration, rather than the original finite-sample optimization
problem.
\par
\endgroup

\end{document}